\documentclass[10pt,twocolumn,letterpaper]{article}
\usepackage[pagenumbers]{cvpr}
\def\confName{CVPR}
\def\confYear{2027}

\usepackage{amsmath,amsfonts,bm}

\def\eqref#1{equation~\ref{#1}}

\def\1{\bm{1}}

\DeclareMathAlphabet{\mathsfit}{\encodingdefault}{\sfdefault}{m}{sl}
\SetMathAlphabet{\mathsfit}{bold}{\encodingdefault}{\sfdefault}{bx}{n}

\usepackage[utf8]{inputenc}
\usepackage[T1]{fontenc}
\usepackage{url}
\usepackage{graphicx}
\usepackage{float}
\usepackage{booktabs}
\usepackage{multirow}
\usepackage{array}
\usepackage{bm}
\usepackage{wrapfig}
\usepackage{pifont}
\usepackage{caption}
\usepackage{amsmath}
\usepackage{amsfonts}
\usepackage{amssymb}
\usepackage{nicefrac}
\usepackage{microtype}
\usepackage{xcolor}
\usepackage{tcolorbox}
\tcbuselibrary{skins,breakable}
\usepackage{fontawesome5}

\definecolor{cvprblue}{rgb}{0.21,0.49,0.74}
\usepackage[pagebackref,breaklinks,colorlinks,allcolors=cvprblue]{hyperref}

\renewcommand{\eg}{e.g.}
\renewcommand{\ie}{i.e.}

\newcommand{\cmark}{\ding{51}}
\newcommand{\xmark}{\ding{55}}

\definecolor{promptbg}{HTML}{F8F9FA}
\definecolor{promptframe}{HTML}{4A90D9}
\definecolor{prompttext}{HTML}{2D3748}

\newtcolorbox{promptbox}[1][]{
  enhanced,
  breakable,
  colback=promptbg,
  colframe=promptframe,
  colbacktitle=promptframe,
  coltitle=promptbg,
  fonttitle=\bfseries\sffamily,
  title={\faRobot\hspace{0.5em}#1},
  arc=2pt,
  boxrule=1.5pt,
  left=10pt, right=10pt, top=2pt, bottom=2pt,
  toptitle=1mm, bottomtitle=1mm,
  shadow={2pt}{-2pt}{0pt}{black!50},
  fontupper=\small\ttfamily\color{prompttext},
}

\title{\texorpdfstring{\raisebox{-0.31\height}{\includegraphics[height=1.8em]{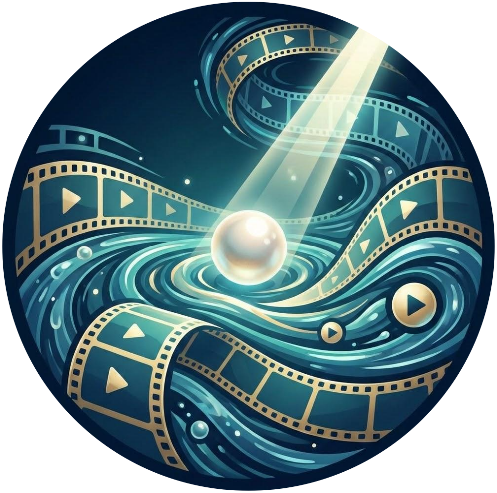}}\hspace{0.15em}}{}PEARL: Personalized Streaming Video Understanding\texorpdfstring{\\}{ } with Runtime-Defined Concepts}

\author{%
Yuanhong Zheng$^{1,*}$\quad
Ruichuan An$^{1,*}$\quad
Xiaopeng Lin$^{1}$\quad
Yuxing Liu$^{1}$\quad
Sihan Yang$^{1}$\\
Huanyu Zhang$^{3}$\quad
Haodong Li$^{4}$\quad
Qintong Zhang$^{1}$\quad
Renrui Zhang$^{5}$\quad
Guopeng Li$^{4}$\\
Yifan Zhang$^{3,\dagger}$\quad
Yuheng Li$^{2,\text{\faEnvelope[regular]}}$\quad
Wentao Zhang$^{1,6,\text{\faEnvelope[regular]}}$\\[0.6em]
{\normalsize
$^{1}$Peking University\quad
$^{2}$Adobe\quad
$^{3}$CASIA\quad
$^{4}$Stepfun\quad
$^{5}$CUHK\quad
$^{6}$Zhongguancun Academy}\\[0.3em]
{\small
$^{*}$Equal contribution\quad
$^{\dagger}$Project leader\quad
$^{\text{\faEnvelope[regular]}}$Corresponding authors}
}

\begin{document}

\maketitle

\begin{abstract}
    Unlike humans, who learn new concepts from a continuous visual stream, existing personalized vision-language models (VLMs) internalize a predefined concept set during training and operate on static images or offline videos. They cannot register a concept that a user defines in the middle of a stream, nor recall the earlier moments in which it appeared. We formalize this task as Personalized Streaming Video Understanding (PSVU) and introduce PSVU-Bench, the first benchmark for this task, which annotates 132 videos with timestamped questions. Its Concept-Definition, Real-Time and Past-Time QA ask a model to register concepts at arbitrary timestamps, recognize their current state, including user-defined actions, and retrieve their historical evidence. We propose PEARL, a strong plug-and-play baseline that stores user-defined concepts apart from archived video clips and rewrites each query with concept descriptions to retrieve the relevant history. Grounding a named concept in video is a skill shared by all concepts. We therefore construct PSVU-10K, a training set of verified concept questions on public internet videos, and train the retriever and answering VLM of PEARL on it once. Concepts defined at test time still require no training. Applied to three architectures without training, PEARL improves Real-Time QA by 15.05\% and Past-Time QA by 12.60\% on average over the corresponding offline models, and all three surpass existing online video models on both QA types. Training further improves PEARL on Qwen3-VL-8B by 8.66\% on Real-Time QA and 9.39\% on Past-Time QA, with gains in every category. Code, data and models are available at \url{https://github.com/Yuanhong-Zheng/PEARL}.
\end{abstract}
\vfill

\begin{figure*}[t]
  \centering
  \includegraphics[width=\linewidth]{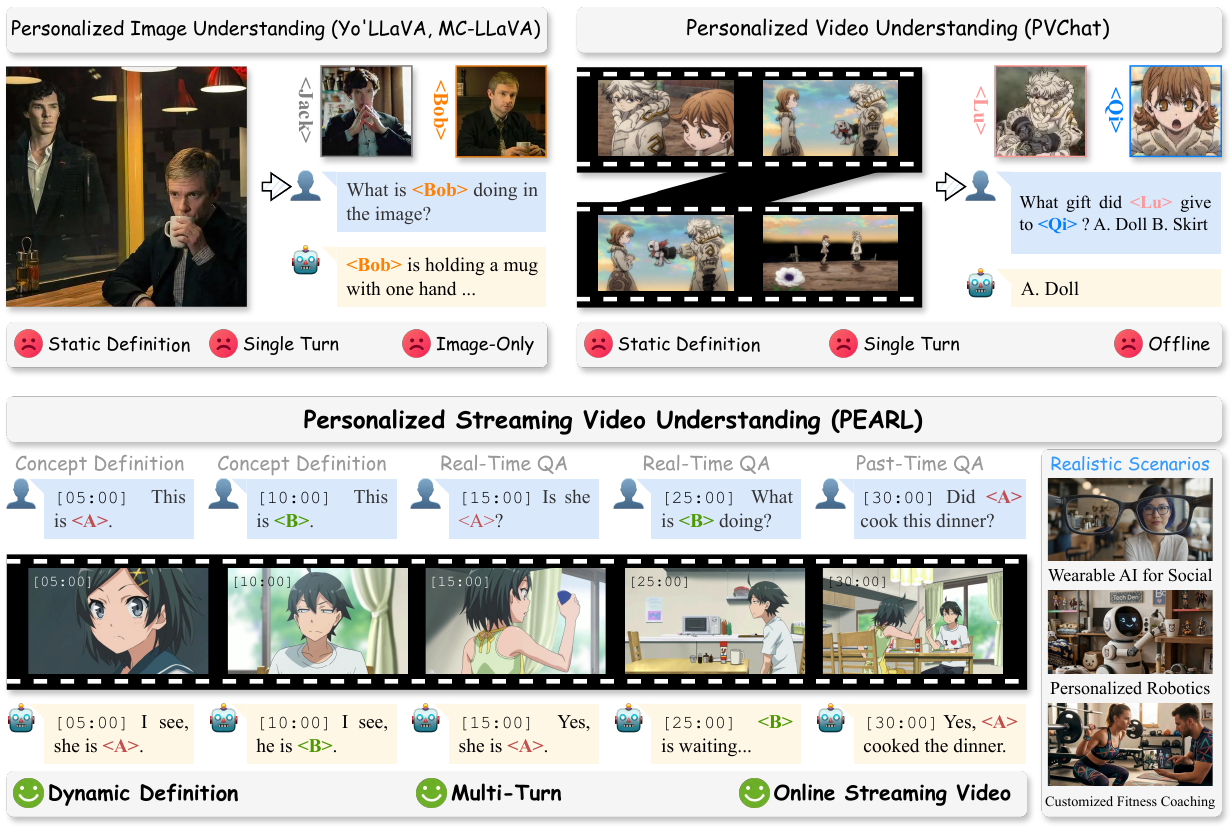} 
  \caption{Comparison between our proposed \textbf{P}ersonalized \textbf{S}treaming \textbf{V}ideo \textbf{U}nderstanding (PSVU) task and traditional Personalized Image/Video Understanding tasks. Unlike traditional settings, PSVU better aligns with real-world scenarios by featuring continuous streaming video inputs, dynamically defined concepts, and multi-turn conversations regarding these customized concepts.}
  \label{fig:Teaser}
  \vspace{-2em}
\end{figure*}

\section{Introduction}
\label{sec:intro}
Imagine a wearable assistant that watches a live video stream with its user. At 05:00 the user says ``This is $\langle$A$\rangle$'', and at 30:00 asks ``Did $\langle$A$\rangle$ cook this dinner?'' (Fig.~\ref{fig:Teaser}). Answering requires binding a name defined during the stream to a person on screen and recalling what that person did since. Personalized vision-language models (VLMs)~\citep{nguyen2406yo,an2024mc,hao2025rap,shi2025pvchat,xu2025jarvis,an2025unictokens,nguyen2025yo,yang2025small,kim2025mmpb} have made clear progress toward such assistants: Yo'LLaVA~\citep{nguyen2406yo} and MC-LLaVA~\citep{an2024mc} learn dedicated tokens for user concepts from a few images, RAP~\citep{hao2025rap} retrieves concept references from a user database, and PVChat~\citep{shi2025pvchat} extends personalization from images to videos.

These methods either learn each concept before deployment or retrieve it from a database of static images, and they operate on images or pre-recorded videos. None of them can register a concept that a user defines in the middle of a stream and later retrieve the moments in which it appeared. Online video models such as ReKV~\citep{di2025streaming}, StreamForest~\citep{zeng2025streamforest} and TimeChat-Online~\citep{yao2025timechat} keep pace with a live stream by caching, merging or dropping visual tokens, but their memory has no link to the names a user defines. Humans handle both demands at once: we learn a name the moment someone is introduced and later recall where we saw that person. Personalized assistants for wearable devices, robots and fitness coaching~\citep{ahmed2025impact,gasteiger2023factors} need the same ability.

We formalize this setting as Personalized Streaming Video Understanding (PSVU) and introduce PSVU-Bench, the first benchmark for it. PSVU-Bench requires \textit{continuous temporal precision}, because every question refers to an exact timestamp of an ongoing stream, and \textit{interactive concept definition}, because users define new concepts at any moment of the stream. Its Concept-Definition QA registers concepts, Real-Time QA queries their current state, and Past-Time QA requires retrieving historical evidence. Real-Time QA also contains an Action category, in which a concept is a user-defined motion grounded in a video clip rather than an entity in a single frame. PSVU-Bench comprises 132 videos and 2,173 annotations with precise timestamps, collected manually or synthesized with AI assistance from anime, movies, reality shows and digital humans.

To address PSVU, we propose PEARL, a plug-and-play framework that equips an off-the-shelf VLM with memory for user-defined concepts. Its Dual-grained Memory System keeps each registered concept, a reference frame or clip with a generated description, apart from a streaming memory that embeds every incoming video clip. Because an embedding model cannot recognize a name that the user has just invented, the Concept-aware Retrieval Algorithm rewrites the query by replacing each concept name with its stored description and uses the rewritten query to retrieve relevant historical clips. The VLM then answers from the current clip, the retrieved clips and the concept references, so a new concept becomes usable as soon as it is defined.

Grounding a named concept in video is a skill shared by all concepts. PEARL therefore learns this skill once before deployment instead of learning each concept. We construct PSVU-10K, a training set of concept questions that an open-source VLM generates and repeatedly verifies on public internet videos. On PSVU-10K we train the query encoder of the retriever while keeping the clip encoder frozen, so the streaming memory never needs to be rebuilt, and we fine-tune the answering VLM. Concepts defined at test time still require no training. No work in PSVU-Bench enters PSVU-10K, so PEARL is always evaluated on concepts it has never trained on. RAP~\citep{hao2025rap} also trains a model once and retrieves user concepts at inference, but it retrieves static concept images, whereas PEARL must also retrieve the moments of a stream in which a concept appeared.

Without training, PEARL raises Real-Time QA by 15.05\% and Past-Time QA by 12.60\% on average when added to LLaVA-OV-7B, Qwen2-VL-7B and Qwen3-VL-8B. With Qwen3-VL-8B, it outperforms all 8 offline and online baselines on both QA types, including the proprietary Gemini 3.1 Pro Preview~\citep{gemini31_2026}. Training on PSVU-10K further improves PEARL on Qwen3-VL-8B by 8.66\% on Real-Time QA and 9.39\% on Past-Time QA, with gains in every category. Action accuracy also rises by 3.90\% although PSVU-10K contains no action concepts, so the skill learned on entities transfers to user-defined motions. Even so, PSVU remains far from solved: the trained model reaches 61.41\% overall against 93.13\% for humans.

Our contributions are as follows:
\begin{itemize}
    \item \textbf{Task and Benchmark:} We define the PSVU task and introduce PSVU-Bench, which evaluates concepts that users define during a video stream.
    \item \textbf{Framework:} We propose PEARL, a strong plug-and-play baseline for PSVU that keeps user-defined concepts apart from the video stream and retrieves historical evidence through concept-aware query rewriting.
    \item \textbf{Training Data:} We construct PSVU-10K, and training PEARL on it once further improves every category, while new concepts still need no training.
\end{itemize}
We review related work on personalized VLMs and streaming video models in Appendix~\ref{app:related}.

\section{PSVU-Bench}
\label{sec:bench}
\subsection{Task Definition}
In PSVU, a model processes a video stream as a continuous sequence of frames. A user can introduce new concepts at any timestamp via instructions (Fig.~\ref{fig:Bench}, left), and for each later query the model must retrieve the relevant concepts and visual context to respond. Unlike personalized VLMs that internalize a predefined concept set and are evaluated on static images or short offline videos~\citep{nguyen2406yo,an2024mc,shi2025pvchat}, PSVU requires concepts introduced at runtime to be maintained throughout the stream. \textbf{Concept-Definition QA} introduces a concept at a specific timestamp, either directly or through context, and the model registers it from the current scene. \textbf{Real-Time QA} asks about registered concepts in the present scene, covering Presence, Behavior, Appearance, Location, Relation and Action. \textbf{Past-Time QA} asks about their earlier states or activities, covering Event-based and Time-based questions, and requires retrieving historical evidence. Concepts can be defined and queried at any later timestamp, so the task is multi-turn. We describe each category in Appendix~\ref{app:qa-categories}.

\subsection{Benchmark Overview}

\begin{figure*}[t]
  \centering
  \includegraphics[width=\linewidth]{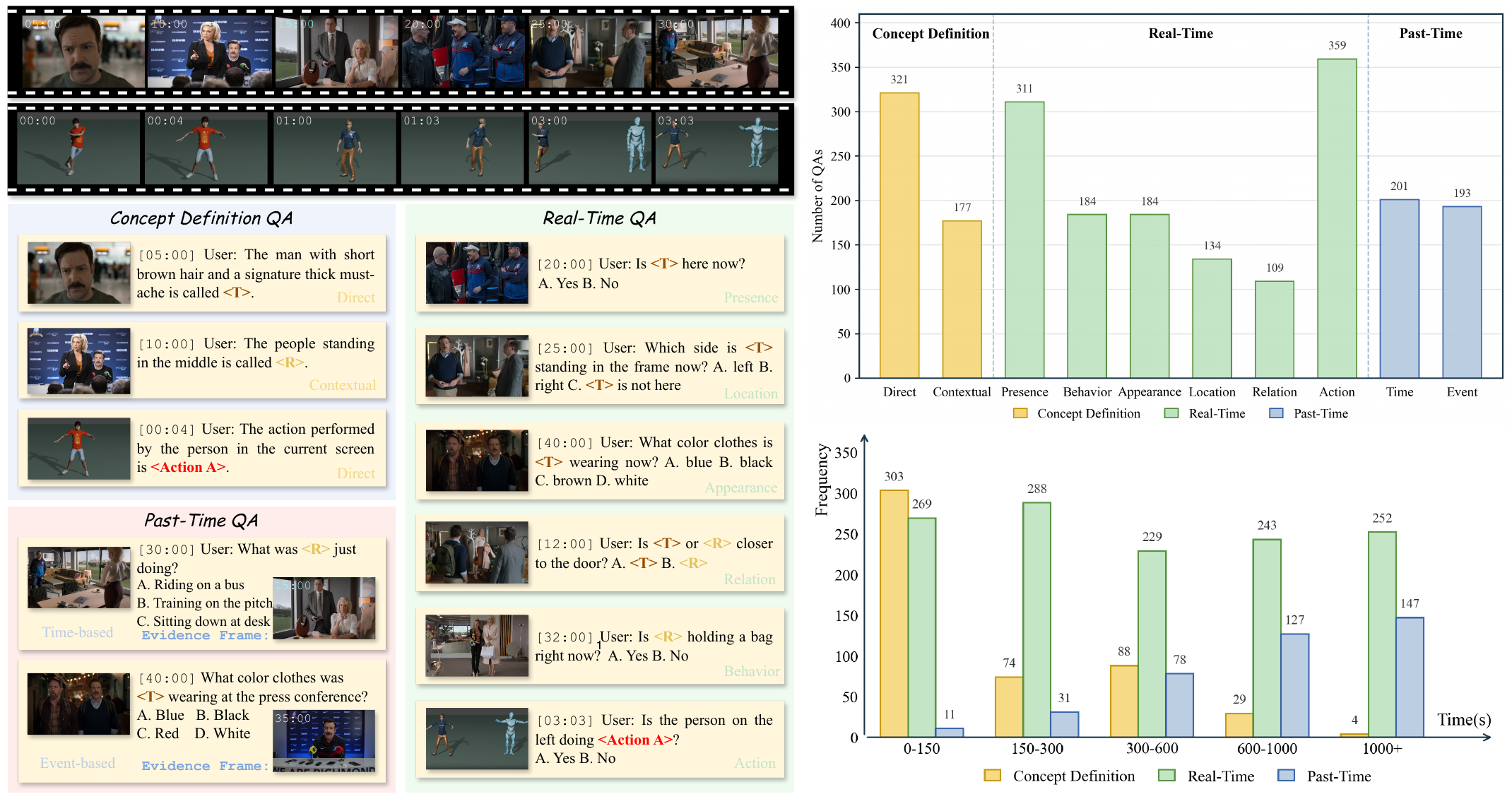} 
  \caption{\textbf{Overview of PSVU-Bench.} \textit{(Left)} A streaming video timeline with representative examples of Concept-Definition QA, Real-Time QA, and Past-Time QA, where Past-Time QA includes explicit historical evidence frames. \textit{(Right)} Benchmark statistics, including the number of QAs across fine-grained categories and the temporal distribution of queries over the video stream.}
  \label{fig:Bench}
\end{figure*}

Table~\ref{tab:bench-comparison} compares PSVU-Bench with existing personalized benchmarks. MyVLM~\citep{alaluf2024myvlm}, Yo'LLaVA~\citep{nguyen2406yo}, MC-LLaVA~\citep{an2024mc}, UnifyBench~\citep{an2025unictokens} and MMPB~\citep{kim2025mmpb} are all image-based, supporting neither video input nor streaming scenarios, and lacking multi-turn interaction. PVChat~\citep{shi2025pvchat} and This-is-My~\citep{yeh2023meta} introduce video modality but are limited to short offline videos (each video is shorter than 5 seconds), with no support for streaming or multi-turn concept interaction. Moreover, none of the above benchmarks supports personalized action concepts, \ie, recognizing user-defined concepts grounded in continuous motions across frames. PSVU-Bench is the first benchmark to simultaneously support long-form streaming video input, multi-turn concept interaction, and personalized concepts covering both entities and actions.
As shown in the right part of Fig.~\ref{fig:Bench}, each QA type in PSVU-Bench contains multiple fine-grained categories. The query timestamps also exhibit a reasonable temporal distribution: Concept-Definition QA is mainly placed in the early stages to define concepts, Real-Time QA is distributed throughout the stream, and Past-Time QA is primarily asked in later stages after sufficient evidence has accumulated.

\begin{table*}[t]
\centering
\caption{\textbf{Comparison of PSVU-Bench with existing personalized benchmarks.}}
\vspace{-3mm}
\label{tab:bench-comparison}
\resizebox{0.78\linewidth}{!}{%
  \begin{tabular}{ccccccc}
  \hline
  \multicolumn{1}{c|}{\multirow{2}{*}{Benchmark}} & \multicolumn{1}{c|}{\multirow{2}{*}{Modality}} & \multicolumn{1}{c|}{\multirow{2}{*}{Streaming}} & \multicolumn{1}{c|}{\multirow{2}{*}{Multi-turn}} & \multicolumn{2}{c|}{Concept Type}             & \multirow{2}{*}{Multi-Concept} \\ \cline{5-6}
  \multicolumn{1}{c|}{}                           & \multicolumn{1}{c|}{}                          & \multicolumn{1}{c|}{}                           & \multicolumn{1}{c|}{}                            & Entity & \multicolumn{1}{c|}{Action} &                                \\ \hline
  MyVLM~\citep{alaluf2024myvlm}                    & Image                                          & --                                              & \xmark & \cmark & \multicolumn{1}{c|}{\xmark} & \xmark                         \\
  Yo'LLaVA~\citep{nguyen2406yo}                    & Image                                          & --                                              & \xmark & \cmark & \multicolumn{1}{c|}{\xmark} & \xmark                         \\
  MC-LLaVA~\citep{an2024mc}                        & Image                                          & --                                              & \xmark & \cmark & \multicolumn{1}{c|}{\xmark} & \cmark                         \\
  UnifyBench~\citep{an2025unictokens}              & Image                                          & --                                              & \xmark & \cmark & \multicolumn{1}{c|}{\xmark} & \xmark                         \\
  MMPB~\citep{kim2025mmpb}                         & Image                                          & --                                              & \xmark & \cmark & \multicolumn{1}{c|}{\xmark} & \cmark                         \\ \hline
  PVChat~\citep{shi2025pvchat}                     & Video (short)                                  & \xmark & \xmark & \cmark & \multicolumn{1}{c|}{\xmark} & \cmark                         \\
  This-is-My~\citep{yeh2023meta}                   & Video (short)                                  & \xmark & \xmark & \cmark & \multicolumn{1}{c|}{\xmark} & \cmark                         \\\hline
  \textbf{PSVU-Bench}                            & Video (long)                                   & \cmark & \cmark & \cmark & \multicolumn{1}{c|}{\cmark} & \cmark                         \\ \hline
  \end{tabular}%
}
\vspace{-1em}
\end{table*}

\subsection{Curation Pipeline}
Our curation pipeline consists of three stages: video collection and filtering, annotation of three QA types (Concept-Definition, Real-Time, and Past-Time), and a quality control phase. We employ a diverse set of question templates to annotate the three QA types. Representative examples are illustrated in Fig.~\ref{fig:Bench}, and Appendix~\ref{app:templates} lists the complete templates.
\subsubsection{Video Collection and Filtering}
We collect videos from YouTube and manually filter them to ensure high dynamics, repeatedly appearing personalized concepts, and resolution no lower than 480p. These videos cover diverse domains including anime, movies, and reality shows. For the Action category, we construct synthetic videos using character models and motions generated with GPT-6 Astra~\citep{openai_gpt6_astra}. Annotators manually arrange and annotate the data, with each distinct action serving as a personalized action concept, as illustrated in Fig.~\ref{fig:Bench}.

\subsubsection{QA Annotation}
For Concept-Definition QA, annotators locate target concepts in the stream and create registration instructions at timestamps where the concept is visible. These items are not included in final scoring, since their role is to introduce user-defined concepts. To reduce reliance on prior knowledge, original concept names are randomly replaced with common names sampled from the U.S.\ SSA database~\citep{ssa_babynames_2026}. For Real-Time QA, annotators identify suitable query timestamps after concept definition and write concept-related questions whose answers require grounding the defined concept in the current scene. For Past-Time QA, annotators additionally specify historical evidence timestamps, ensuring that each question cannot be answered from the current clip alone and instead requires retrieving and reasoning over earlier stream content. Representative examples are shown in the left part of Fig.~\ref{fig:Bench}, and Appendix~\ref{app:visualization} provides more examples.

\subsubsection{Quality Control}
To ensure high annotation quality, our curation team consists of 10 researchers with multimodal research experience, including 6 annotators and 4 reviewers. We adopt a combined pipeline of automated filtering and human verification. In the automated stage, we apply an ablation-based filtering method similar to Sec.~\ref{subsubsec:effectiveness}: Real-Time QA is tested with and without provided concepts, while Past-Time QA is tested with and without historical evidence clips. Questions that remain answerable when the required information is withheld are deemed trivial and filtered. In the human verification stage, reviewers conduct multiple rounds of manual inspection to ensure that each QA item and timestamp is aligned with the video content. We additionally collect human evaluation scores as an upper-bound reference, reported in Table~\ref{tab:main-table}. Appendix~\ref{app:annotation-quality} gives more annotation and quality-control details.

\vspace{-0.5em}
\section{PEARL Framework}
\label{sec:pearl-framework}
\begin{figure*}[t]
  \centering
  \includegraphics[width=\linewidth]{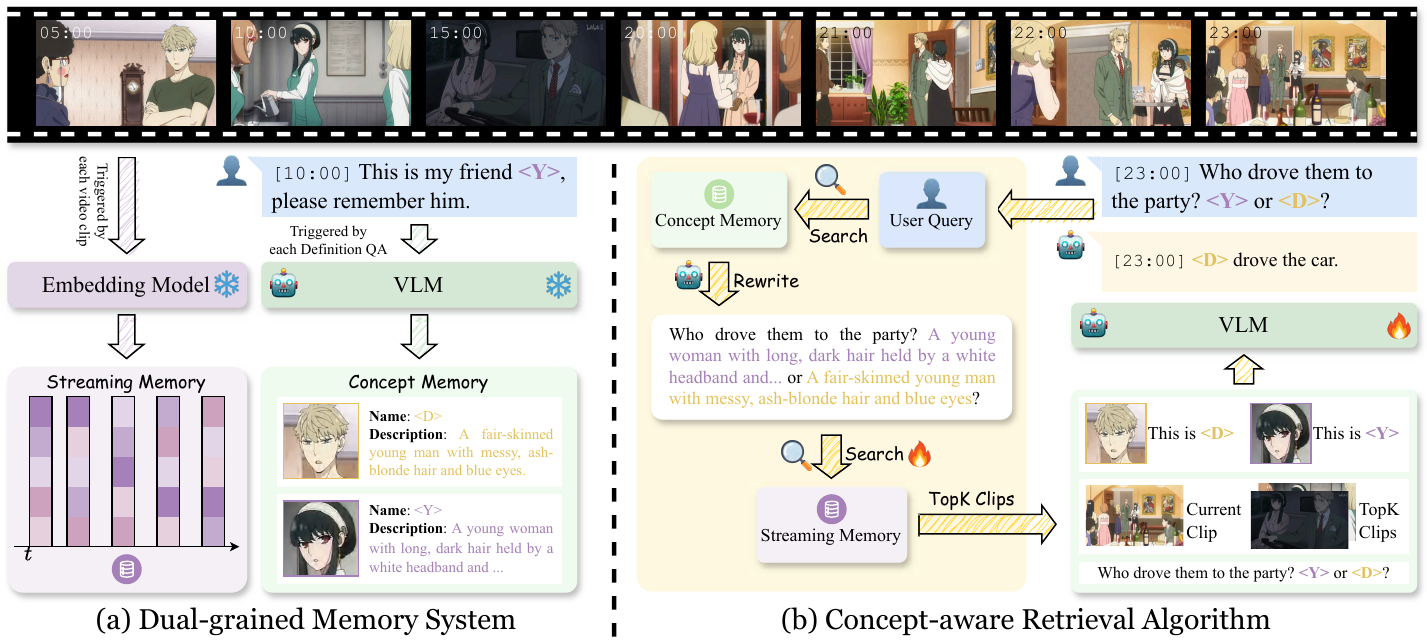} 
  \caption{\textbf{PEARL framework.} \textbf{(a) Dual-grained Memory System:} \textit{Concept Memory} stores user-defined concepts with visual evidence and textual descriptions, while \textit{Streaming Memory} archives segmented clips with multimodal embeddings. \textbf{(b) Concept-aware Retrieval Algorithm:} Upon a user query, PEARL retrieves relevant concepts and top-$K$ historical clips, then feeds them together with the current clip into a VLM for real-time personalized response. Flame icons mark the modules trained on PSVU-10K (query encoder and answering VLM), and snowflake icons mark frozen ones (clip encoder and the VLM that writes concept descriptions).}
  \label{fig:Algorithm}
  \vspace{-1.5em}
\end{figure*}

PEARL (Fig.~\ref{fig:Algorithm}) registers concepts from user instructions at any timestamp of a stream and answers later queries in real time. After formulating the task (Sec.~\ref{subsec:problem-formulation}), we describe its Dual-grained Memory System (Sec.~\ref{subsec:dual-grained-memory}) and Concept-aware Retrieval Algorithm (Sec.~\ref{subsec:concept-aware-retrieval}), which work with off-the-shelf models, and then train its retriever and answering VLM once on PSVU-10K (Sec.~\ref{subsec:training}).

\subsection{Formulation}
\label{subsec:problem-formulation}
Formally, we define a streaming video as an infinite sequence $V = [\mathcal{X}_1, \mathcal{X}_2, \dots]$, where $\mathcal{X}_i$ denotes a video clip representing a semantic scene. Throughout the stream, a user can introduce new concepts at any timestamp $t_c$ via instructions, forming an evolving set of defined concepts $\mathcal{C} = \{C_1, C_2, \dots\}$. For a query $Q$ issued at time $t_q \geq t_c$, the model $\mathcal{M}$ must dynamically construct a context to generate a response $A$:
\begin{equation}
    A = \mathcal{M}(\mathcal{C}_{\mathrm{sub}}, \mathcal{V}_{\mathrm{context}}, Q)
\end{equation}
where $\mathcal{C}_{\mathrm{sub}} \subseteq \mathcal{C}$ is the query-relevant concept subset, and $\mathcal{V}_{\mathrm{context}}$ is the necessary visual context. Solving this requires overcoming two key challenges: the prohibitive cost of maintaining unbounded stream history alongside evolving concepts, and the difficulty of accurately retrieving personalized $\mathcal{C}_{\mathrm{sub}}$ and $\mathcal{V}_{\mathrm{context}}$ in real time. This motivates our design of a scalable dual-grained memory and a concept-aware retrieval strategy.

\vspace{-0.5em}
\subsection{Dual-grained Memory System}
\label{subsec:dual-grained-memory}
PSVU requires a model to retain concepts defined at arbitrary timestamps and to access long-range evidence from the stream. The Dual-grained Memory System therefore keeps a Streaming Memory of embedded clips apart from a Concept Memory of user-defined concepts.

\subsubsection{Streaming Memory}
Streaming Memory maintains a set of entries, each consisting of a video clip $\mathcal{X}_i$ and its corresponding embedding $\mathbf{e}_i$. Given a continuously arriving video stream, we first detect scene boundaries and segment the stream into an ordered sequence of clips $V = [\mathcal{X}_1, \mathcal{X}_2, \dots]$. For each newly detected clip $\mathcal{X}_i$, we employ a multimodal embedding model $f_{\mathrm{emb}}(\cdot)$ to compute an embedding $\mathbf{e}_i = f_{\mathrm{emb}}(\mathcal{X}_i)$, and store the pair $(\mathcal{X}_i, \mathbf{e}_i)$ in Streaming Memory. Each clip embedding $\mathbf{e}_i$ captures semantic information about the scene and is used for retrieval. Detailed settings are provided in Appendix~\ref{app:impl}.

\subsubsection{Concept Memory}

When a Concept-Definition query $Q_{\mathrm{def}}$ is issued at timestamp $t_c$, the model creates a new entry with three components: (i) a concept name, (ii) associated visual evidence, and (iii) a textual description. The model first invokes an external tool to extract the visual evidence from the current clip $\mathcal{X}_{t_c}$, using either the most relevant frame or the corresponding short clip depending on the concept-definition instruction. Conditioned on this extracted visual evidence, the model then generates a compact description that summarizes the concept’s salient characteristics, using a standardized prompting template provided in Appendix~\ref{app:prompts}. The resulting entry is finally inserted into the Concept Memory for subsequent retrieval and querying.

\subsection{Concept-aware Retrieval Algorithm}
\label{subsec:concept-aware-retrieval}
When a user issues a Real-Time query or a Past-Time query at timestamp $t_q$, the model needs three types of information to respond accurately: (i) the query $Q$, (ii) the query-relevant concept subset $\mathcal{C}_{\mathrm{sub}}\subseteq\mathcal{C}$ retrieved from Concept Memory, and (iii) the visual evidence $\mathcal{V}_{\mathrm{context}}$ retrieved from Streaming Memory, producing the final answer as $A=\mathcal{M}(\mathcal{C}_{\mathrm{sub}},\mathcal{V}_{\mathrm{context}},Q)$. To obtain $\mathcal{C}_{\mathrm{sub}}$, we identify the concept names mentioned in $Q$ and use them as keys to retrieve the corresponding Concept Memory entries. To obtain $\mathcal{V}_{\mathrm{context}}$, we use the model $\mathcal{M}$ to rewrite the query into $\tilde{Q}$ by replacing each concept name with its associated description, then encode it using the same multimodal embedding model as Streaming Memory to compute $\mathbf{e}_Q=f_{\mathrm{emb}}(\tilde{Q})$. We compute cosine similarities between $\mathbf{e}_Q$ and all stored clip embeddings $\{\mathbf{e}_i\}_{i\le t_q}$ (where $\mathbf{e}_i=f_{\mathrm{emb}}(\mathcal{X}_i)$), select the top-$K$ most relevant clips, and further expand each selected clip with its adjacent $N$ clips to capture temporally local context, yielding $\mathcal{V}_{\mathrm{context}}\subseteq\{\mathcal{X}_1,\dots,\mathcal{X}_{t_q}\}$. Finally, we feed the retrieved concept entries, the retrieved historical clips, the current clip $\mathcal{X}_{t_q}$, and the original query $Q$ into the VLM to generate the response, preserving real-time responsiveness while maximizing retrieval of task-relevant evidence.

\subsection{Training PEARL}
\label{subsec:training}
Without training, PEARL relies on general-purpose models for two learned steps. The embedding model has never learned to match a rewritten query to the clips in which a described concept appears, and the VLM has never learned to answer from concept references together with retrieved clips that may include distractors. Both abilities are shared by all concepts, so we learn them once on a dedicated training set instead of learning each new concept.

\noindent\textbf{PSVU-10K.} We build PSVU-10K from YouTube videos. Each sample is a causal stream prefix of up to 15 minutes, split into short clips, with two to four concepts that carry randomly sampled names, a reference frame and a description. At the end of the prefix, a four-option Real-Time or Past-Time question asks about these concepts, and its supporting evidence clips are annotated. An open-source VLM writes each question and then verifies it: the VLM must select the annotated answer uniquely after reviewing the whole prefix without any hint of the evidence location, must answer correctly from the evidence clips and concept references alone, and must fail once either the concept definition or the evidence is withheld. The verifier also confirms up to two clips that do not support the answer, which serve as negatives. PSVU-10K contains 10k questions (Appendix~\ref{app:psvu10k}), split by source video into 9k for training and 1k for validation. We use the work name in each video's metadata to exclude every PSVU-Bench work from PSVU-10K, so PEARL never trains on the concepts it is evaluated on.

\noindent\textbf{Retriever.} We adapt only the query path of the embedding model, denoted $f_{\mathrm{emb}}^{\theta}$, and keep $f_{\mathrm{emb}}$ frozen for clips, so the embeddings already stored in Streaming Memory remain valid. For the rewritten query $\tilde{Q}$ produced by the inference pipeline, let $s_j$ denote the similarity between $f_{\mathrm{emb}}^{\theta}(\tilde{Q})$ and the embedding $\mathbf{e}_j$ of clip $j$. With the evidence clips $\mathcal{P}$ and the verified negatives $\mathcal{N}$, we minimize
\begin{equation}
    \mathcal{L}_{\mathrm{ret}} = -\frac{1}{|\mathcal{P}|}\sum_{p\in\mathcal{P}}\log\frac{\exp(s_p/\tau)}{\sum_{j\in\mathcal{P}\cup\mathcal{N}}\exp(s_j/\tau)},
    \label{eq:retrieval-loss}
\end{equation}
where $\tau$ is a temperature. Averaging over all evidence clips rewards the retriever for covering every piece of evidence rather than only the easiest one. Clips that the verifier has not examined are never used as negatives, because they may contain unlabeled evidence.

\noindent\textbf{Answering VLM.} We fine-tune $\mathcal{M}$ to output the correct option given the concept references, the current clip, the evidence clips mixed with the verified negatives, and the question with its options, using the standard cross-entropy loss on the answer. Each question appears under all four option rotations, which discourages position bias.

\section{Experiments}
\label{sec:exp}

\subsection{Implementation Details}
\label{subsec:impl-details}

As a plug-and-play framework, we employ Qwen3-VL-Embedding-2B~\citep{li2026qwen3} as the multimodal embedding model, and use LLaVA-OV-7B~\citep{li2024llava}, Qwen2-VL-7B~\citep{wang2024qwen2}, and Qwen3-VL-8B~\citep{bai2025qwen3} as base models, respectively. For the Dual-grained Memory System, we use the corresponding base model to generate concept descriptions and rewrite user queries. We use PySceneDetect~\citep{castellano_pyscenedetect} to detect scene boundaries and segment the streaming video into clips. For the Concept-aware Retrieval Algorithm, we set $K=4$, $N=1$, and sample the input video stream at 1 FPS.

As a pioneering effort in this area, where no directly comparable baselines exist, we instead adopt representative offline and online video understanding methods as our baselines. For offline models, we adopt different sampling protocols according to the queried concept type: for entity-related questions, we uniformly sample 64 frames, and for action-related questions, we use a 64-second window and sample at 1 FPS to ensure the model observes continuous motion patterns. For online models, given the unique paradigm of the PSVU task, which necessitates concept registration prior to question answering, we exclusively select models capable of multi-turn dialogue as our baselines. Furthermore, we strictly follow their original sampling configurations. The rationale for these settings is discussed in Appendix~\ref{app:baseline-settings}. All experiments are conducted on NVIDIA H200 GPUs. Note that to minimize the model's bias toward answer options and prevent random guessing~\citep{pezeshkpour2024large,atabuzzaman2025benchmarking}, we evaluate each question using a cyclic option rotation strategy (Appendix~\ref{app:option-rotation}).

For training, we apply LoRA~\citep{hu2022lora} to both the retriever and the answering VLM of Qwen3-VL-8B+PEARL and fine-tune them on PSVU-10K, with hyperparameters in Appendix~\ref{app:training-settings}. We evaluate the trained PEARL on the full PSVU-Bench and compare training configurations on its 112 entity videos.

\begin{table*}[t]
\vspace{-1em}
\centering
\caption{\textbf{Results on PSVU-Bench.} We report fine-grained Real-Time QA and Past-Time QA metrics with their respective averages, where Overall Avg denotes the average of Real-Time QA Avg and Past-Time QA Avg. $^\dagger$ denotes training on PSVU-10K, and all other models are used without training. \textbf{Bold} and \underline{underline} denote the best and second-best results among open-source models and PEARL, respectively.}
\label{tab:main-table}
\setlength{\tabcolsep}{4pt}
\resizebox{\textwidth}{!}{%
\begin{tabular}{@{} l c c|ccccccc|ccc|c @{}}
\toprule
\multirow{2}{*}{Method} & \multirow{2}{*}{\#Frames} & \multirow{2}{*}{\shortstack{Token\\Budget}} & \multicolumn{7}{c|}{Real-Time QA} & \multicolumn{3}{c|}{Past-Time QA} & \multirow{2}{*}{Overall Avg} \\ \cmidrule(lr){4-10} \cmidrule(lr){11-13}
& & & Presence & Behavior & Appearance & Location & Relation & Action & Avg & Event & Time & Avg & \\ \midrule
\multicolumn{14}{c}{Human Score} \\ \midrule
Human & - & -- & 94.21 & 93.48 & 94.02 & 93.28 & 94.50 & 93.59 & 93.85 & 93.26 & 91.54 & 92.40 & 93.13 \\ \midrule
\multicolumn{14}{c}{Text-only} \\ \midrule
Qwen3-VL-8B~\citep{bai2025qwen3} & - & 0 & 12.54 & 10.87 & 11.41 & 9.70 & 11.01 & 6.96 & 10.42 & 19.17 & 15.42 & 17.30 & 13.86 \\ \midrule
\multicolumn{14}{c}{Proprietary Offline Model} \\ \midrule
Gemini 3.1 Pro Preview~\citep{gemini31_2026} & 64 & -- & 50.16 & 46.74 & 48.91 & 44.78 & 45.87 & 24.51 & 43.50 & 51.81 & 46.27 & 49.04 & 46.27 \\ \midrule
\multicolumn{14}{c}{Open-source Offline Model} \\ \midrule
LLaVA-OV-7B~\citep{li2024llava} & 64 & 12,544 & 27.01 & 23.37 & 25.00 & 23.13 & 25.69 & 10.86 & 22.51 & 36.27 & 31.84 & 34.06 & 28.29 \\
Qwen2-VL-7B~\citep{wang2024qwen2} & 64 & 24,576 & 25.08 & 21.74 & 22.83 & 22.39 & 23.85 & 17.83 & 22.29 & 38.34 & 33.33 & 35.84 & 29.07 \\
InternVL3.5-8B~\citep{wang2025internvl3} & 64 & 16,384 & 32.15 & 29.35 & 31.52 & 28.36 & 28.44 & 5.57 & 25.90 & 37.31 & 32.84 & 35.08 & 30.49 \\
Qwen3-VL-8B~\citep{bai2025qwen3} & 64 & 24,576 & 28.94 & 26.09 & 27.72 & 26.12 & 28.44 & 25.63 & 27.16 & 32.64 & 27.86 & 30.25 & 28.71 \\ \midrule
\multicolumn{14}{c}{Open-source Online Model} \\ \midrule
ReKV (LLaVA-OV-7B)~\citep{di2025streaming} & 0.5fps & 12,544 & 27.97 & 25.00 & 26.63 & 23.88 & 27.52 & 24.23 & 25.87 & 39.38 & 35.32 & 37.35 & 31.61 \\
StreamForest-7B~\citep{zeng2025streamforest} & 1fps & 8,192 & 31.19 & 28.80 & 29.89 & 27.61 & 28.44 & 10.86 & 26.13 & 43.01 & 38.31 & 40.66 & 33.40 \\
TimeChat-Online-7B~\citep{yao2025timechat} & 1fps & 6,000 & 34.08 & 30.43 & 32.07 & 30.60 & 31.19 & 22.28 & 30.11 & 37.31 & 33.33 & 35.32 & 32.72 \\ \midrule
\multicolumn{14}{c}{PEARL Framework} \\ \midrule
\textbf{LLaVA-OV-7B+PEARL} & 1fps & 8,192 & 35.05 & 32.61 & 34.24 & 31.34 & 34.86 & 20.06 & 31.36 & 45.08 & 40.30 & 42.69 & 37.03 \\
\textbf{Qwen2-VL-7B+PEARL} & 1fps & 8,192 & 35.37 & 32.07 & 33.70 & 31.34 & 33.94 & 24.23 & 31.78 & 47.15 & 41.79 & 44.47 & 38.13 \\
\textbf{Qwen3-VL-8B+PEARL} & 1fps & 8,192 & \underline{63.67} & \underline{50.00} & \underline{55.43} & \underline{52.99} & \underline{53.21} & \underline{48.47} & \underline{53.96} & \underline{52.85} & \underline{48.76} & \underline{50.80} & \underline{52.38} \\
\textbf{Qwen3-VL-8B+PEARL}$^\dagger$ & 1fps & 8,192 & \textbf{68.49} & \textbf{62.50} & \textbf{62.50} & \textbf{65.67} & \textbf{64.22} & \textbf{52.37} & \textbf{62.62} & \textbf{62.18} & \textbf{58.21} & \textbf{60.19} & \textbf{61.41} \\ \bottomrule
\end{tabular}%
}
\vspace{-1em}
\end{table*}
\subsection{Main Results}
We report Human Score and a Text-only baseline as upper and lower bounds in Table~\ref{tab:main-table}. The large gap between them shows that PSVU-Bench is highly solvable with sufficient visual information, but cannot be reliably answered from text priors alone and requires grounding in the streaming video. The comparisons below use PEARL without training, and Sec.~\ref{subsec:training-results} analyzes the trained PEARL, marked with $^\dagger$.

\subsubsection{Comparison with Offline Baselines}
We evaluate several representative open-source offline models, along with a strong proprietary offline baseline. As shown in Table~\ref{tab:main-table}, these offline approaches struggle on the PSVU task. In contrast, applying PEARL to the same base models yields consistent and substantial improvements. On Real-Time QA, LLaVA-OV-7B+PEARL, Qwen2-VL-7B+PEARL, and Qwen3-VL-8B+PEARL improve their offline counterparts by 8.85\%, 9.49\%, and 26.80\%, respectively. On Past-Time QA, the corresponding gains are 8.63\%, 8.63\%, and 20.55\%. Qwen3-VL-8B+PEARL also exceeds the proprietary Gemini 3.1 Pro Preview by 6.11\% in Overall Avg. Action is the hardest category for every offline model, and PEARL raises it by 22.84\% on Qwen3-VL-8B, since the reference clip shows the motion that the user names. These results demonstrate the generality and robustness of PEARL across distinct architectures. The performance gap reveals a key limitation of offline models in the streaming setting: to bound inference latency and token cost, they operate on a fixed set of sampled frames and typically lack explicit memory mechanisms to preserve and retrieve long-range historical evidence. Consequently, the evidence for many queries never reaches the model.

\subsubsection{Comparison with Online Baselines}
We compare PEARL against three representative open-source online models: ReKV (LLaVA-OV-7B)~\citep{di2025streaming}, StreamForest-7B~\citep{zeng2025streamforest}, and TimeChat-Online-7B~\citep{yao2025timechat}. As shown in Table~\ref{tab:main-table}, PEARL consistently outperforms these online baselines on both Real-Time QA and Past-Time QA. Compared with the strongest online baseline on each QA type, Qwen3-VL-8B+PEARL achieves gains of 23.85\% on Real-Time QA and 10.14\% on Past-Time QA. Notably, LLaVA-OV-7B+PEARL comprehensively outperforms ReKV on both metrics, achieving gains of 5.49\% and 5.34\%, respectively. Since ReKV shares the same backbone and is likewise a plug-and-play framework, this controlled comparison shows that the gains stem from the design of PEARL rather than the backbone. Online models compress history into a fixed-size state with no link to user-defined names, whereas PEARL stores each concept and retrieves the clips relevant to it.

\subsection{Effect of Training}
\label{subsec:training-results}
\begin{table*}[t]
\centering
\caption{\textbf{Effect of training on PSVU-Bench.} PEARL with Qwen3-VL-8B on the entity concepts of PSVU-Bench, with the retriever and the answering VLM either kept frozen or trained with LoRA on PSVU-10K. Real-Time and Past-Time averages are means over categories. Recall@$k$ counts evaluable Past-Time questions whose annotated evidence is among the top-$k$ retrieved clips, and Hit also includes the adjacent clips that PEARL feeds to the VLM.}
\label{tab:training}
\setlength{\tabcolsep}{4pt}
\resizebox{\textwidth}{!}{%
\begin{tabular}{@{} cc|cccccc|ccc|ccc @{}}
\toprule
\multicolumn{2}{c|}{Trained} & \multicolumn{6}{c|}{Real-Time QA} & \multicolumn{3}{c|}{Past-Time QA} & \multicolumn{3}{c}{Retrieval} \\ \cmidrule(lr){1-2} \cmidrule(lr){3-8} \cmidrule(lr){9-11} \cmidrule(lr){12-14}
Retriever & VLM & Presence & Behavior & Appearance & Location & Relation & Avg & Event & Time & Avg & R@1 & R@4 & Hit \\ \midrule
\xmark & \xmark & 63.67 & 50.00 & 55.43 & 52.99 & 53.21 & 55.06 & 52.85 & 48.76 & 50.80 & 7.85 & 22.25 & 34.82 \\
\cmark & \xmark & 63.02 & 50.54 & 54.89 & 51.49 & 52.29 & 54.45 & 46.63 & 45.27 & 45.95 & \textbf{14.66} & \textbf{32.46} & \textbf{44.50} \\
\xmark & \cmark & 66.24 & \textbf{62.50} & 61.96 & 63.43 & \textbf{64.22} & 63.67 & 60.10 & 55.72 & 57.91 & 7.85 & 22.25 & 34.82 \\
\cmark & \cmark & \textbf{68.49} & \textbf{62.50} & \textbf{62.50} & \textbf{65.67} & \textbf{64.22} & \textbf{64.68} & \textbf{62.18} & \textbf{58.21} & \textbf{60.19} & \textbf{14.66} & \textbf{32.46} & \textbf{44.50} \\ \bottomrule
\end{tabular}%
}
\vspace{-1em}
\end{table*}
On the full PSVU-Bench, training both the retriever and the answering VLM on PSVU-10K improves Qwen3-VL-8B+PEARL by 8.66\% on Real-Time QA and 9.39\% on Past-Time QA, with gains in every category (Table~\ref{tab:main-table}). Location and Behavior gain the most, by 12.68\% and 12.50\%. Action accuracy also rises by 3.90\% although PSVU-10K contains no action concepts, which indicates that the learned grounding extends beyond the concept types seen in training. Training brings similar gains on LLaVA-OV-7B and Qwen2-VL-7B (Appendix~\ref{app:train-other-vlms}).

Table~\ref{tab:training} compares training configurations on the entity concepts. Training only the answering VLM brings most of the gain, 8.61\% on Real-Time QA and 7.11\% on Past-Time QA, as it learns to ground registered concepts in the current clip and to answer from retrieved evidence. Training the retriever raises Recall@1 by 6.81\% and the evidence hit rate of the expanded context by 9.68\%. With the original VLM, however, this retrieval lowers Past-Time accuracy by 4.85\%. We attribute this drop to clips that show the queried concept without answering the question. The trained retriever ranks such clips higher along with the evidence, and the original VLM is misled by them. The trained VLM has learned to ignore them from the verified negatives of PSVU-10K and reaches the best Past-Time accuracy of 60.19\% with the same retrieval. Training the retriever therefore pays off only together with a trained VLM.

\subsection{Ablation Study}
Training leaves the inference pipeline of PEARL unchanged, so unless stated otherwise, the following analyses use PEARL without training. This setting isolates the contribution of the framework design and matches the plug-and-play use of PEARL with any VLM.
\subsubsection{Effectiveness of PEARL Design}
\label{subsubsec:effectiveness}

We ablate PEARL by progressively enabling its components in Table~\ref{tab:my-table}. Without visual input, Qwen3-VL-8B answers only 13.86\% of the questions, and the current clip alone adds little. Concept Memory raises Real-Time accuracy by 35.57\%, because the model can link a user-defined name to its entity only through the stored concept. Streaming Memory then raises Past-Time accuracy by 20.26\%, since these questions require historical clips rather than the current scene. Query Rewriting adds a further 4.42\% to the average, as the embedding model matches the descriptions that replace concept names but not the names themselves.

On evaluable frame-level Past-Time questions, PEARL's Top-4 retrieval with adjacent-clip expansion includes the annotated evidence timestamp in 34.82\% of cases, compared with a 7.90\% expected hit rate for uniform random selection with the same expansion (Appendix~\ref{app:evidence-retrieval}).

Real-world users may define a concept late or ambiguously, and buffering and selective clarification extend PEARL to these cases (Appendix~\ref{app:interaction-robustness}). On the non-personalized OVO-Bench~\citep{niu2025ovo}, PEARL also raises the average accuracy of LLaVA-OV-7B by 2.60\% and Qwen2-VL-7B by 2.55\%, so it adds personalization without harming general streaming understanding (Appendix~\ref{app:general-streaming}).

\begin{table}[htbp]
\vspace{-1em}
\centering
\caption{\textbf{Ablation of PEARL components} on PSVU-Bench with Qwen3-VL-8B, showing progressive gains from Text-only to the full PEARL pipeline.}
\label{tab:my-table}
\resizebox{\linewidth}{!}{%
\begin{tabular}{@{}ccccc|ccc@{}}
\toprule
Text & Current & Concept & Streaming & Rewrite & Real-Time & Past-Time & Avg \\ \midrule
\cmark & \xmark  & \xmark  & \xmark   & \xmark  & 10.42          & 17.30          & 13.86 \\
\cmark & \cmark  & \xmark  & \xmark   & \xmark  & 15.84          & 20.30          & 18.07 \\
\cmark & \cmark  & \cmark  & \xmark   & \xmark  & 51.41          & 25.43          & 38.42 \\
\cmark & \cmark  & \cmark  & \cmark   & \xmark  & 50.22          & 45.69          & 47.96 \\
\cmark & \cmark  & \cmark  & \cmark   & \cmark  & \textbf{53.96} & \textbf{50.80} & \textbf{52.38} \\ \bottomrule
\end{tabular}%
}
\end{table}

\subsubsection{Efficiency of PEARL Design} 

\begin{table}[t]
\centering
\caption{Comparison of end-to-end inference latency across models. Overall Avg denotes the average of Real-Time QA Avg and Past-Time QA Avg.}
\label{tab:eff-latency-right}
\resizebox{\linewidth}{!}{%
\begin{tabular}{@{}lccc@{}}
\toprule
Method & \#Frames & Overall Avg & Latency (ms) \\ \midrule
LLaVA-OV-7B & 64 & 28.29 & 670 \\
Qwen3-VL-8B & 64 & 28.71 & 1,594 \\ \midrule
ReKV (LLaVA-OV-7B) & 0.5fps & 31.61 & 1,818 \\
StreamForest-7B & 1fps & 33.40 & 1,164 \\
TimeChat-Online-7B & 1fps & 32.72 & 4,769 \\ \midrule
\textbf{LLaVA-OV-7B+PEARL} & 1fps & 37.03 & 775 \\
\textbf{Qwen3-VL-8B+PEARL} & 1fps & {52.38} & 2,111 \\ \bottomrule
\end{tabular}%
}
\end{table}

We further evaluate end-to-end inference efficiency. As shown in Table~\ref{tab:eff-latency-right}, PEARL adds a modest latency overhead to its base models, 105 ms for LLaVA-OV-7B and 517 ms for Qwen3-VL-8B, while raising their Overall Avg by 8.74\% and 23.67\%. LLaVA-OV-7B+PEARL is both more accurate and faster than all online baselines, and Qwen3-VL-8B+PEARL outperforms the strongest online baseline, StreamForest-7B, by 18.98\% while remaining faster than TimeChat-Online-7B. The VLM input stays bounded as the stream grows, because PEARL passes at most 12 retrieved clips and the current clip to the VLM.

Concept retrieval, query rewriting and streaming memory retrieval together take about 200 ms for both models, so LLM inference accounts for most of the latency. Appendix~\ref{app:online-streaming} breaks the latency down by stage and describes concurrent streaming inference and its overhead.

\section{Conclusion}
\label{sec:conclusion}
We introduced Personalized Streaming Video Understanding, in which users define concepts at any moment of a video stream and later ask about their present and past, and built PSVU-Bench to evaluate it. As a strong plug-and-play baseline, PEARL keeps user-defined concepts apart from the stream and retrieves historical evidence through concept-aware query rewriting, which improves three VLM architectures without training. Training PEARL once on PSVU-10K further improves every category, including user-defined actions absent from PSVU-10K, while new concepts still require no training. These gains hold for all three VLM architectures. Personalization in streaming video can thus be learned once as a general skill, so an assistant handles each new concept the moment a user names it.

\section*{LLM Usage Statement}
The research ideas, task formulation, method design, experimental protocols, and interpretation of results were developed by the authors. We used AI tools as follows.

\noindent\textbf{Writing assistance.} We used GPT-6.1 Sol to polish text drafted by the authors, including grammar, word choice, and fluency, and to format references.

\noindent\textbf{Benchmark construction.} We used GPT-6 Astra to generate character models and motions for the synthetic Action videos, which were manually arranged and annotated. Human annotators determined the query timestamps, questions, and correct answers. We used Gemini 3.1 Pro Preview to generate multiple-choice distractors for Real-Time and Past-Time QA from these human-specified questions and answers and the corresponding video clips.

\noindent\textbf{Code implementation.} We used GPT-6.1 Sol to assist with writing, completing, and debugging code, including the Dual-grained Memory System and the Concept-aware Retrieval Algorithm, according to algorithm designs and component specifications provided by the authors.

All AI outputs in these three uses were verified by the authors, and revised or discarded where necessary. The authors take full responsibility for the paper, code, and benchmark.

\noindent\textbf{Automated screening.} We used Gemini 3.1 Pro Preview for the ablation-based filtering described in Appendix~\ref{app:annotation-quality}.

\noindent\textbf{Training data construction.} We used GPT-6 Luna to rank candidate source videos from their metadata, and the open-source Qwen3.8-Flash-Next to generate and verify the PSVU-10K questions automatically (Appendix~\ref{app:psvu10k}). These questions are used only for training and validation and are not part of PSVU-Bench.

\noindent\textbf{Controlled interaction study and appendix writing.} A Codex assistant helped develop the clarification prompt, prepare visual reference expressions, and assess image--text concept bindings. The visual assessments were not independently reviewed by human annotators. GPT-6.1 Terra drafted the text on this study and the annotation procedure.

\renewcommand{\bibsection}{\section*{References}}
{
\small
\bibliographystyle{ieeenat_fullname}
\bibliography{main}
}

\clearpage
\onecolumn
\appendix

\section{Related Works}
\label{app:related}
\noindent\textbf{Personalized VLMs.} As VLMs continue to advance~\citep{wang2025internvl3,bai2025qwen3,li2024llava,yu2025minicpm,gemini31_2026,zhang2024long,zhang2024llava,gpt4v_2023,liu2025nvila}, growing attention has turned to using them as personalized assistants~\citep{hong2025dialogue,cohen2022my,wu2024personalized, oh2026contextualized}. Most work personalizes image understanding, through fine-tuning~\citep{alaluf2024myvlm,nguyen2406yo,an2024mc,yang2025small}, retrieval-augmented generation~\citep{hao2025rap,xu2025jarvis} or reinforcement learning~\citep{oh2025repic,feng2026m2a}. These methods operate on static images and do not extend to video. Other studies unify personalized understanding and generation~\citep{nguyen2025yo,an2025unictokens,zhong2026unified}, but they also rely on concepts defined before inference. In video, early work~\citep{yeh2023meta} focuses on personalized retrieval, and PVChat~\citep{shi2025pvchat} introduces personalized video question answering for offline videos. RAP~\citep{hao2025rap} is closest to our training design, because it trains a model once and retrieves user concepts at inference. PEARL follows the same principle but must also retrieve the moments of a stream in which a concept appeared.

\noindent\textbf{Streaming video understanding.} Streaming video models~\citep{di2025streaming, yao2025timechat, zeng2025streamforest, niu2025ovo, yang2025svbench, xun2025rtv, lin2024streamingbench, chen2024videollm, qian2025dispider, fu2025vita} process continuous visual input for real-time interaction, but they do not support concepts that users define during the stream. No existing setting therefore combines streaming input, real-time responses and flexible concept definition. We introduce the PSVU task and PSVU-Bench for this setting, and propose PEARL as a strong plug-and-play baseline that registers concepts and answers in real time within a video stream.

\section{QA Sub-categories}
\label{app:qa-categories}
To thoroughly evaluate model capabilities under diverse scenarios, we further categorize the queries in PSVU-Bench. We systematically classify the questions within Concept-Definition QA, Real-Time QA, and Past-Time QA into fine-grained sub-categories based on their reasoning requirements.

\noindent\textbf{Concept-Definition QA.} We classify the concept definitions into two distinct types:
\begin{itemize}
    \item \textbf{Direct:} The concept is defined straightforwardly without relying on complex descriptions. For entity concepts, it points out the main character or object in the scene (e.g., \textit{``This character is called \{ConceptName\}. Please remember this name.''}). For action concepts, it identifies a specific personalized motion pattern over a continuous clip (e.g., \textit{``The sequence of movements shown in this clip is defined as \{ConceptName\}. Please remember this name.''}).
    \item \textbf{Contextual:} The concept is defined by describing its explicit visual attributes (like clothing color) or its interactions and relationships with other objects in the scene (e.g., \textit{``The character wearing white clothes is named \{ConceptName\}. Please remember this name.''}).
\end{itemize}

\noindent\textbf{Real-Time QA.} To evaluate multi-dimensional perception, we categorize the real-time queries into six distinct sub-tasks:
\begin{itemize}
    \item \textbf{Presence:} Asking whether the concept is present in the current scene (e.g., \textit{``Is \{ConceptName\} here now?''}).
    \item \textbf{Behavior:} Querying the current action or state of the defined concept (e.g., \textit{``What is \{ConceptName\} doing now?''}).
    \item \textbf{Appearance:} Focusing on the transient visual details of the concept, such as current clothing (e.g., \textit{``What color is \{ConceptName\} wearing now?''}).
    \item \textbf{Location:} Identifying the position of the concept in the scene (e.g., \textit{``Where is \{ConceptName\} located in this scene now?''}).
    \item \textbf{Relation:} Inquiring about the interaction or relationship between the target concept and other entities or objects (e.g., \textit{``Who is standing next to \{ConceptName\} now?''}).
    \item \textbf{Action:} Querying whether a dynamically defined action concept is being performed across the current continuous clip (e.g., \textit{``Is the person doing \{ConceptName\} now?''}).
\end{itemize}

\noindent\textbf{Past-Time QA.} We divide the past-time queries into two types based on the retrieval mechanism required to answer them:
\begin{itemize}
    \item \textbf{Event-based:} The historical evidence can be localized purely based on a specific event or action, without requiring strict temporal reasoning (e.g., \textit{``What was \{ConceptName\} holding when he was cooking?''}).
    \item \textbf{Time-based:} The historical evidence requires the model to understand the temporal sequence or order of events to accurately retrieve the correct clip (e.g., \textit{``What did \{ConceptName\} do right before he left the room?''}).
\end{itemize}

\section{Complete Question Templates}
\label{app:templates}
As mentioned in the main text, we provide the complete question templates used in PSVU-Bench for annotating Concept-Definition QA, Real-Time QA, and Past-Time QA. In Table~\ref{tab:question_templates}, we summarize the representative question templates corresponding to each fine-grained sub-category. It is worth noting that these question templates serve primarily as a reference for the annotators. In practice, annotators often diversify the content, phrasing, and sentence structures of the questions based on specific video scenarios, thereby ensuring the richness and naturalness of the data. To represent the varied entities and properties systematically, we utilize the following placeholders in the templates:
\begin{itemize}
    \item \texttt{\{ConceptName\}}: The dynamically assigned user-defined name for the target concept.
    \item \texttt{\{OtherConcept\}}: The name of another previously defined concept.
    \item \texttt{\{EntityPronoun\}}: General references to the subject, such as \textit{person, character, man, girl, object}, etc.
    \item \texttt{\{ConceptFeature\}}: Specific visual characteristics or attributes used to describe the concept, such as \textit{short hair, red clothes, sitting by the window}, etc.
    \item \texttt{\{Object\}}: Variables pertaining to the question, representing objects interacted with or queried.
    \item \texttt{\{Location\}}: A specific location or scene setting referenced in the queries.
    \item \texttt{\{Event\}}: A specific action or event referenced in the queries.
    \item \texttt{\{DefinitionSuffix\}}: A standardized suffix appended to concept definitions to enforce model compliance. For entity concepts, it is: \textit{``From now on, regardless of who this character might be in real life or in any media, you must refer to them as \{ConceptName\}.''} For action concepts, it is: \textit{``From now on, regardless of how this action might be typically described or what similar movements exist, whenever you see this specific pattern of motion, you must refer to it as \{ConceptName\}.''}
\end{itemize}

\begin{table}[t]
\centering
\caption{Representative question templates across different QA types and sub-categories in PSVU-Bench.}
\label{tab:question_templates}
\resizebox{\textwidth}{!}{%
\begin{tabular}{@{}llp{11.5cm}@{}}
\toprule
\textbf{QA Type} & \textbf{Sub-category} & \textbf{Question Templates} \\ \midrule
\multirow{4}{*}{Concept-Definition} & Direct & \textbullet~This \{EntityPronoun\} is called \{ConceptName\}. Please remember this name. \{DefinitionSuffix\} \newline \textbullet~This \{EntityPronoun\} is my friend, named \{ConceptName\}. Please remember this name. \{DefinitionSuffix\} \newline \textbullet~The sequence of movements shown in this clip is defined as \{ConceptName\}. Please remember this name. \{DefinitionSuffix\}  \\ \cmidrule(l){2-3} 
 & Contextual & \textbullet~The \{EntityPronoun\} in the \{Location\} is named \{ConceptName\}. Please remember this name. \{DefinitionSuffix\} \newline \textbullet~The \{EntityPronoun\} with \{ConceptFeature\} in the scene is my friend, named \{ConceptName\}. Please remember this name. \{DefinitionSuffix\} \newline \textbullet~The \{EntityPronoun\} who is \{Event\} is called \{ConceptName\}. Please remember this name. \{DefinitionSuffix\} \\ \midrule
\multirow{12}{*}{Real-Time} & Presence & \textbullet~Is this \{EntityPronoun\} \{ConceptName\} at this moment? \newline \textbullet~Are \{ConceptName\} and \{OtherConcept\} here now? \newline \textbullet~Is \{ConceptName\} currently among this group of people? \\ \cmidrule(l){2-3} 
 & Behavior & \textbullet~What is \{ConceptName\} doing now? \newline \textbullet~Is \{ConceptName\} \{Event\} right now? \newline \textbullet~What is \{ConceptName\}'s expression now? \\ \cmidrule(l){2-3} 
 & Appearance & \textbullet~What color is \{ConceptName\} wearing now? \newline \textbullet~What is the shape of the \{Object\} \{ConceptName\} is wearing? \newline \textbullet~What color is \{ConceptName\}'s \{Object\} now? \\ \cmidrule(l){2-3} 
 & Location & \textbullet~Where is \{ConceptName\} located in the frame? \newline \textbullet~Is \{ConceptName\} on the \{Location\} in the picture? \newline \textbullet~Where is \{ConceptName\} now? \\ \cmidrule(l){2-3} 
 & Relation & \textbullet~Who is standing next to \{ConceptName\} now? \newline \textbullet~What is \{ConceptName\} giving to \{OtherConcept\} right now? \newline \textbullet~What is \{ConceptName\} holding right now? \\ \cmidrule(l){2-3} 
 & Action & \textbullet~Is the person doing \{ConceptName\} now? \newline \textbullet~What is the person doing? Is it \{ConceptName\}? \\ \midrule
\multirow{4}{*}{Past-Time} & Event-based & \textbullet~Did \{ConceptName\} just walk into the \{Location\}? \newline \textbullet~What color clothes was \{ConceptName\} wearing when \{Event\}? \newline \textbullet~What did \{ConceptName\} just put into the \{Object\}? \\ \cmidrule(l){2-3} 
 & Time-based & \textbullet~Is \{ConceptName\}'s hair color the same as it was just now? \newline \textbullet~What was \{ConceptName\} holding before \{EntityPronoun\} tried to \{Event\}? \newline \textbullet~What was \{ConceptName\} doing before \{OtherConcept\} \{Event\}? \\ \bottomrule
\end{tabular}%
}
\end{table}

\section{Prompt Templates}
\label{app:prompts}

\subsection{Concept Description Generation Prompts}
We provide standardized prompting templates used to generate a compact description that summarizes the concept's salient characteristics. Since PSVU-Bench evaluates both entity and action concepts, we design a separate prompt for each type:

\noindent\textbf{Entity Concept Prompt.} This prompt is designed to guide the model to focus on permanent and stable visual features (such as gender, facial features, hair, and body type) while instructing it to ignore temporary elements like clothing, accessories, or poses, which are likely to change across a long video stream.

\noindent\textbf{Action Concept Prompt.} Conversely, this prompt directs the model to focus on the core kinematics and stable movement patterns of a customized action, while explicitly ignoring the specific identity or appearance of the person performing it, as well as the background and surrounding environment, so that the description generalizes across characters and scenes.

Specifically, both templates include two key placeholders:
\begin{itemize}
    \item \texttt{\{concept\_name\}}: The user-defined name assigned to the concept.
    \item \texttt{\{original\_description\}}: The user instruction from the Concept-Definition QA. For entity concepts, it helps the model locate the target subject (e.g., ``The character wearing white clothes is named \{Adaliz\}.''). For action concepts, it helps the model identify the specific action (e.g., ``The sequence of movements shown in this clip is \{Action A\}.'').
\end{itemize}

The complete prompts for both concept types are provided below:

\begin{promptbox}[Entity Concept Description Generation Prompt]
Based on the image and the original description provided, generate a concise visual description of this character/object that focuses on PERMANENT/STABLE features for video clip retrieval.\\[0.5em]
Original description: "\{original\_description\}"\\
Concept name: \{concept\_name\}\\[0.5em]
Your task:\\
1. Use the original description to understand WHICH character/object to focus on in the image\\
2. Generate a description focusing on STABLE features that DON'T change throughout the video:\\
\hspace*{1em} - Gender (male/female/other)\\
\hspace*{1em} - Face features (eye shape, facial structure, distinctive marks)\\
\hspace*{1em} - Hair (color, length, style if distinctive)\\
\hspace*{1em} - Body type (build)\\
\hspace*{1em} - Age appearance (young/middle-aged/elderly)\\[0.5em]
AVOID or minimize:\\
- Clothing details (they change in long videos)\\
- Accessories (they may be removed)\\
- Temporary expressions or poses\\
- Background, location, surroundings, or nearby objects in the scene\\
- Relative position or size compared to objects/environment in the scene\\[0.5em]
Requirements:\\
- Keep it concise and simple (1 sentence, around 10-15 words)\\
- Focus on features that remain consistent across different scenes\\
- Write in English using simple descriptive terms\\
- Use third person (e.g., "a young male with...", "the girl with...")\\
- Make it natural enough to replace the concept name in a question\\[0.5em]
Please provide the distinctive visual description focusing on PERMANENT features:
\end{promptbox}

\begin{promptbox}[Action Concept Description Generation Prompt]
Based on the provided video clip and the original description, generate a concise textual description of the specific ACTION or MOVEMENT PATTERN that focuses on the CORE KINEMATICS for video clip retrieval.\\[0.5em]
Original description: "\{original\_description\}"\\
Concept name: \{concept\_name\}\\[0.5em]
Your task:\\
1. Use the original description to understand WHICH specific action or sequence of movements to focus on in the video clip\\
2. Generate a description focusing on the STABLE MOVEMENT PATTERNS that define this action, regardless of who is performing it:\\
\hspace*{1em} - Core body movements (e.g., raising arms, squatting, twisting)\\
\hspace*{1em} - Sequence of motions (the order of the gestures)\\
\hspace*{1em} - Body parts involved (hands, legs, torso)\\
AVOID or minimize:\\
- The specific identity, gender, age, or appearance of the person performing the action\\
- Background, location, surroundings, or irrelevant objects in the scene\\
- Any static features that do not contribute to the dynamic action itself\\[0.5em]
Requirements:\\
- Keep it concise and simple (1 sentence, around 10-20 words)\\
- Focus strictly on the dynamic movement pattern that can be performed by different characters\\
- Write in English using simple descriptive action terms\\
- Use general action phrases (e.g., "the action of swinging arms in a circle", "the action of squatting down and then leaping forward")\\
- Make it natural enough to replace the concept name in a question\\[0.5em]
Please provide the distinctive action description focusing on CORE MOVEMENT PATTERNS:
\end{promptbox}

\subsection{Query Rewrite Prompt}
As described in Sec.~\ref{subsec:concept-aware-retrieval}, we use a prompt to rewrite user queries by replacing concept names with their descriptions to improve retrieval accuracy. This process helps translate user-defined concept names (which the multimodal embedding model has not seen) into explicit visual or kinematic semantics that facilitate accurate historical clip retrieval.

Specifically, the template includes two key placeholders:
\begin{itemize}
    \item \texttt{\{query\}}: The original user question containing the customized concept names.
    \item \texttt{\{replacement\_instructions\}}: A set of automatically constructed rules mapping each concept name found in the query to its generated description (e.g., \textit{``\{Adaliz\}'' becomes ``a young female with long black hair''} for an entity concept, or \textit{``\{Action A\}'' becomes ``the action of squatting down and then leaping forward''} for an action concept).
\end{itemize}

The complete prompt is provided below:

\begin{promptbox}[Query Rewrite Prompt]
Rewrite the following question by replacing the concept names (in curly braces) with their visual descriptions. Keep the sentence grammatically correct and natural.\\[0.5em]
Original question:\\
\{query\}\\[0.5em]
Replacement rules:\\
\{replacement\_instructions\}\\[0.5em]
Requirements:\\
- Replace each \{ConceptName\} with the provided visual description\\
- Adjust grammar as needed (e.g., articles, verb forms) to keep the sentence natural\\
- Do NOT change the meaning of the question\\
- Do NOT add or remove any information\\
- Output ONLY the rewritten question, nothing else
\end{promptbox}

\subsection{Question Answering Prompt}
\label{app:answer-prompt}
The answering VLM receives a single interleaved message. Each concept referenced in the question contributes its reference frame, or its reference clip for an action concept, followed by its \texttt{\{original\_description\}}. The retrieved historical clips and the current clip follow, and the message ends with the question, its four options and an instruction to answer in \texttt{<ans>} tags. All PEARL variants are evaluated with this format, and training on PSVU-10K uses the same format. The complete prompt is provided below, where bracketed italic entries denote visual inputs:

\begin{promptbox}[Question Answering Prompt]
\textit{[Reference frame or clip of each concept]}\\
\{original\_description\}\\[0.5em]
Here are the previous related video clips:\\
\textit{[Retrieved historical clips]}\\[0.5em]
The following is the current video clip:\\
\textit{[Current clip]}\\[0.5em]
\{question\}\\
A. \{option\_A\}\\
B. \{option\_B\}\\
C. \{option\_C\}\\
D. \{option\_D\}\\[0.5em]
Please output your answer choice in <ans></ans> tags.
\end{promptbox}

\section{More Visualization Examples}
\label{app:visualization}
To better illustrate the complexity and diversity of our dataset, as well as the effectiveness of our proposed framework, we present additional visualization examples from PSVU-Bench. These examples demonstrate the model's capability to handle continuous video streams, dynamically register personalized concepts, and answer both real-time and past-time queries.

\begin{figure}[htbp]
  \centering
  \includegraphics[width=\linewidth]{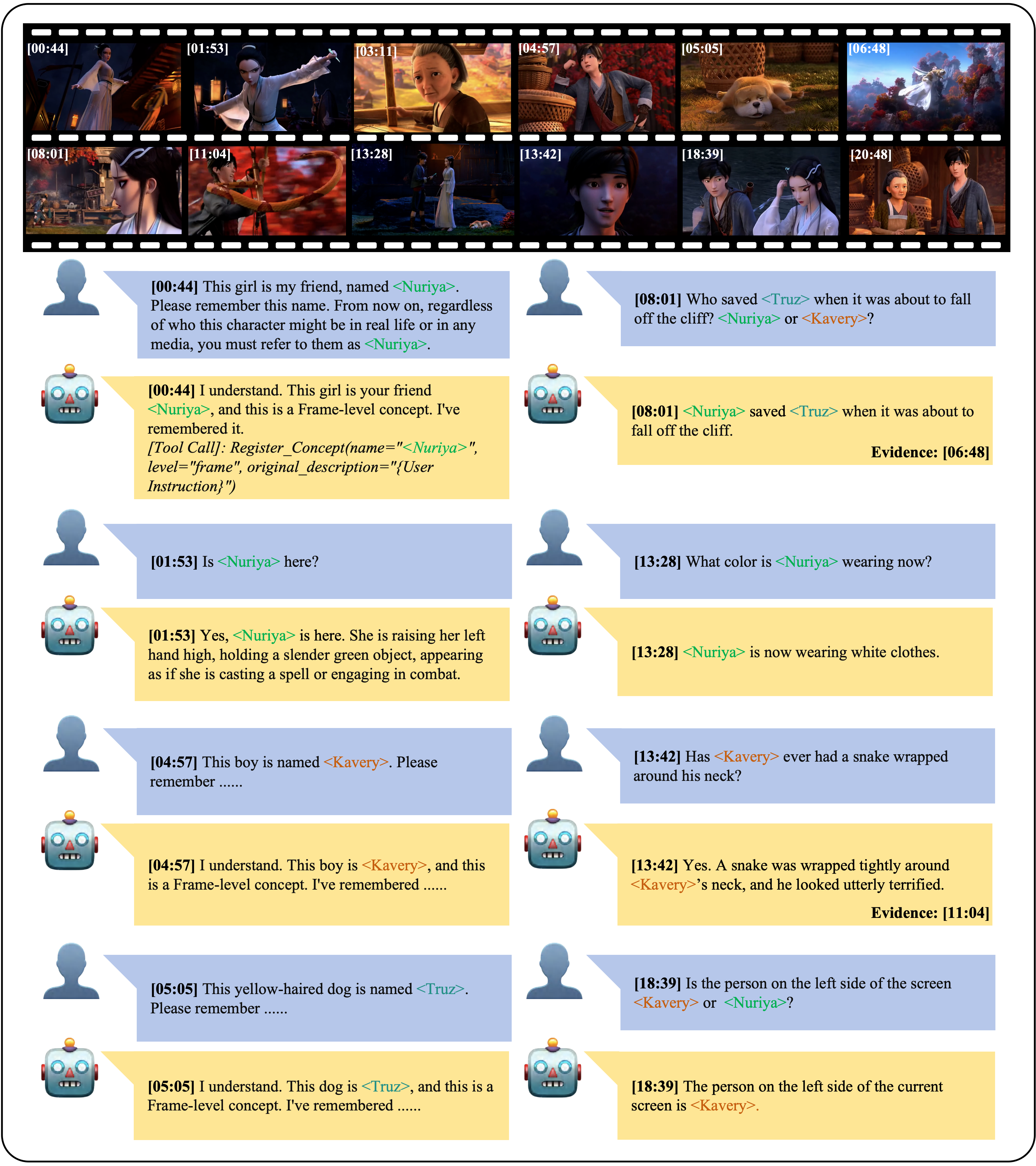} 
  \caption{Visualization example of multi-turn interactions in PSVU-Bench. The model registers multiple user-defined concepts (\eg, \texttt{<Nuriya>}, \texttt{<Kavery>}, \texttt{<Truz>}) and answers subsequent Real-Time and Past-Time queries by retrieving the corresponding historical evidence.}
  \label{fig:appendix_01}
\end{figure}

As shown in Fig.~\ref{fig:appendix_01}, we showcase a comprehensive multi-turn interaction process using the Qwen3-VL-8B+PEARL model on a long animated video. As the video stream progresses, the user dynamically defines multiple concepts at different timestamps (\eg, \texttt{<Nuriya>} at [00:44], \texttt{<Kavery>} at [04:57], and \texttt{<Truz>} at [05:05]). Upon receiving these Concept-Definition instructions, the PEARL framework successfully invokes the \texttt{Register\_Concept} tool. This tool specifically functions to extract the current visual evidence (\ie, the current frame), generate a concise visual description for the target concept, and subsequently update the Concept Memory by storing the visual evidence, the generated description, and the concept name as a unified entity.

Subsequently, the model demonstrates robust real-time perception by accurately answering Real-Time QA based on the current scene. For example, at [01:53], the model not only successfully recognizes \texttt{<Nuriya>}'s presence but also accurately describes her combat posture. Later, at [18:39], it correctly identifies \texttt{<Kavery>} on the left side of the screen. Furthermore, the model exhibits strong long-term temporal reasoning in Past-Time QA. For instance, at [08:01], when asked who saved \texttt{<Truz>}, the model successfully retrieves the historical evidence from [06:48] to provide the correct answer. Similarly, at [13:42], it accurately recalls the terrifying snake encounter that occurred at [11:04]. These results highlight PEARL's ability to maintain and retrieve personalized memories effectively.

\begin{figure}[t]
  \centering
  \includegraphics[width=\linewidth]{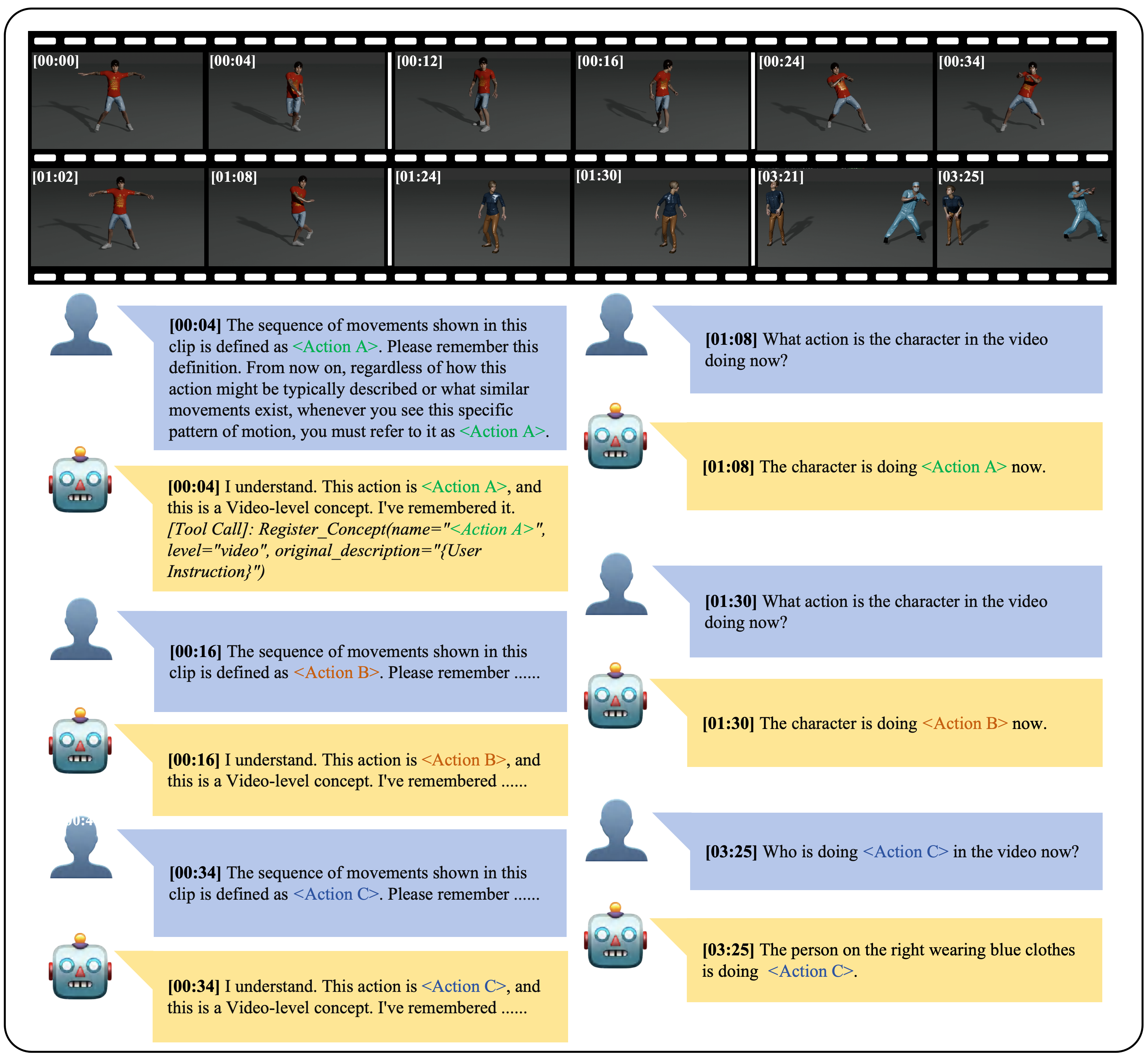} 
  \caption{Additional visualization example for the Action category. The model registers multiple user-defined action sequences (\eg, \texttt{<Action A>}, \texttt{<Action B>}, \texttt{<Action C>}) and recognizes these customized actions when they are performed by different characters later in the video stream.}
  \label{fig:appendix_02}
\end{figure}

We also provide an additional example for the Action category in Fig.~\ref{fig:appendix_02}. The model registers several user-defined action sequences (\eg, \texttt{<Action A>}, \texttt{<Action B>}, and \texttt{<Action C>}) and recognizes them when they reappear later in the stream, showing that PEARL can also support personalized motion concepts.

\section{Detailed Implementation Settings}
\label{app:impl}

\subsection{Multimodal Embedding Model and Scene Detection}
We provide additional details regarding the multimodal embedding model and scene detection configurations used in our Dual-grained Memory System.

\noindent\textbf{Multimodal Embedding Model.} We adopt Qwen3-VL-Embedding-2B~\citep{li2026qwen3} as our multimodal embedding model. This model encodes the rewritten queries and the streaming video clips into a shared embedding space, and historical clips are retrieved by their cosine similarity to the query. Video clips are sampled at 1 FPS for embedding extraction. 

\noindent\textbf{Scene Detection.} We employ PySceneDetect~\citep{castellano_pyscenedetect} to segment the continuous streaming video into semantically coherent clips by detecting fast cuts based on pixel changes in the HSV colorspace between adjacent frames. We set the scene detection threshold to 27.0. To ensure the segmented clips contain sufficient temporal context while avoiding excessively long segments that might dilute the semantic focus, we enforce a minimum clip duration of 1.0 second and a maximum clip duration of 8.0 seconds. Any detected scene longer than the maximum duration is split into consecutive 8-second segments.

\subsection{Online Streaming Inference}
\label{app:online-streaming}
\begin{figure}[htbp]
\centering
\includegraphics[width=0.55\linewidth]{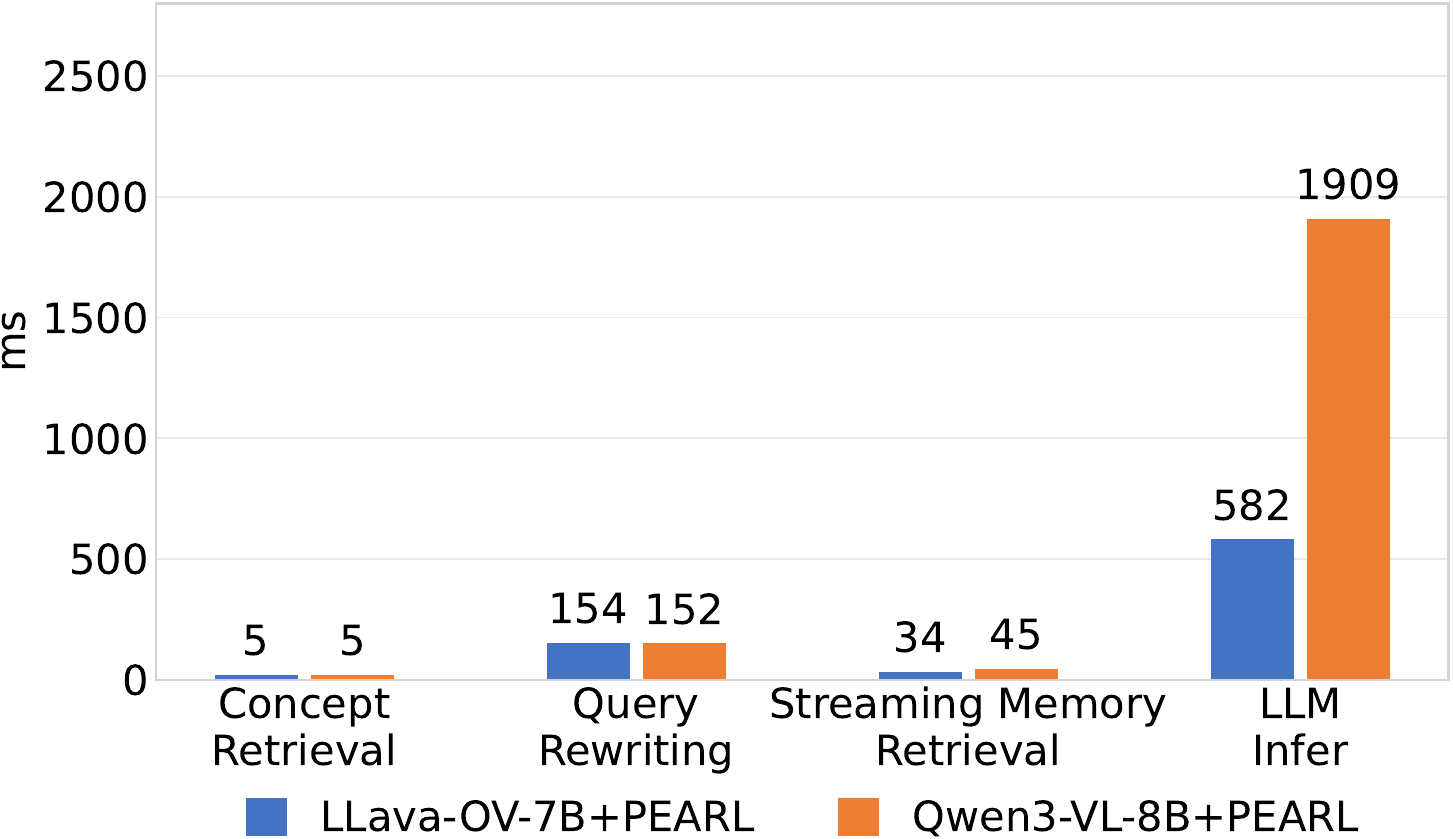}
\caption{End-to-end latency breakdown of PEARL across different models.}
\label{fig:Hyperparameter2}
\end{figure}

Fig.~\ref{fig:Hyperparameter2} breaks the end-to-end latency in Table~\ref{tab:eff-latency-right} down by stage. Concept retrieval, query rewriting and streaming memory retrieval take 193 ms for LLaVA-OV-7B and 202 ms for Qwen3-VL-8B, while LLM inference takes 582 ms and 1,909 ms, respectively.

We implement a causal streaming prototype with separate workers for video reception and scene detection, clip encoding and memory insertion, and user interaction. As frames arrive, the first worker applies the scene detector incrementally. It closes a clip at a detected boundary once at least one second of video has accumulated, or at eight seconds if no eligible boundary has occurred. The remaining partial clip is flushed at the end of the stream. A background worker materializes and embeds each completed clip while video reception continues. Concept registrations and questions are processed by a separate worker. When serving a question, the model uses the buffered current clip up to the question time and historical clips that ended by that time and whose embeddings are available. No frames after the question time are used. This causal splitting procedure can produce different clips from the offline preprocessing used for the main benchmark.

We measure the CPU stages on a 60-second, 854$\times$480 video replayed at 24 FPS on a dual-socket Intel Xeon 6776P host. Frame decoding and online scene detection take 1.19 and 3.76 ms per frame on average, respectively. Materializing the 17 completed clips with FFmpeg takes 0.21 s per clip on average. In a separate 210-second replay with 56 pre-segmented clips, Qwen3-VL-Embedding-2B on an NVIDIA H200 processes a clip in 0.36 s on average from clip arrival to memory insertion, while the VLM runs on a second H200. Both measurements use a recorded video, without camera or network input.

The question-to-answer latency in the concurrent deployment is approximately 2.11 s, while the interval from a completed clip boundary to its insertion into memory is approximately 0.57 s.

\subsection{Cyclic Option Rotation Evaluation Strategy}
\label{app:option-rotation}
We employ a cyclic option rotation strategy during evaluation. Questions with fewer than four substantive options are padded with \texttt{None} entries to obtain four option slots before evaluation. Thus, every question is evaluated four times, with its correct answer placed once at each position (A, B, C, and D), regardless of its original number of substantive options. Each variation swaps the originally correct option with the target option while leaving all other positions unchanged.

For question $i$, let $\hat{a}_i^{(r)}$ and $a_i^{(r)}$ denote the predicted and ground-truth option labels in variation $r$, respectively. The question-level score is
\begin{equation}
    s_i = \prod_{r=1}^{4} \mathbf{1}\!\left[\hat{a}_i^{(r)} = a_i^{(r)}\right],
    \label{eq:rotation-score}
\end{equation}
where $\mathbf{1}[\cdot]$ is the indicator function. A question receives a score of one only when all four variations are answered correctly, and any incorrect variation makes its score zero. Questions originally containing two or three options therefore follow the same four-variation scoring rule after padding, rather than being evaluated only two or three times.

For a fine-grained QA category $c$ with question set $\mathcal{Q}_c$, its accuracy is
\begin{equation}
    \operatorname{Acc}_c = \frac{100}{|\mathcal{Q}_c|} \sum_{i\in\mathcal{Q}_c} s_i.
    \label{eq:rotation-accuracy}
\end{equation}
We average the category accuracies within Real-Time QA and Past-Time QA separately, and report their mean as Overall Avg in Table~\ref{tab:main-table}. This metric measures correctness consistently across answer positions and is stricter than accuracy evaluated under a single option ordering.

\subsection{Baseline Settings}
\label{app:baseline-settings}
We evaluate the online baselines at their original sampling rates, 0.5 FPS for ReKV~\citep{di2025streaming} and 1 FPS for StreamForest~\citep{zeng2025streamforest} and TimeChat-Online~\citep{yao2025timechat}, while PEARL processes the stream at 1 FPS. Each rate reflects how the method balances temporal coverage against computation, and raising it would also increase the visual input and inference latency of that baseline. We therefore keep the original rates so that each baseline is evaluated at its intended operating point.

\subsection{Training Settings}
\label{app:training-settings}
We train the query path of Qwen3-VL-Embedding-2B and the answering Qwen3-VL-8B with LoRA~\citep{hu2022lora} of rank 16, $\alpha=32$ and dropout 0.05. Both are optimized with AdamW, a weight decay of 0.01, gradient clipping at 1.0 and an effective batch size of 8. The retriever is trained on the Past-Time questions of PSVU-10K for 2,244 steps with a learning rate of $2\times10^{-5}$ and a contrastive temperature of $\tau=0.07$. The VLM is trained for 9,140 steps, about two epochs over the four option rotations, with a learning rate of $10^{-4}$. Inputs are capped at 8,192 tokens. LLaVA-OV-7B and Qwen2-VL-7B (Sec.~\ref{app:train-other-vlms}) reuse the trained retriever, and their answering VLMs are trained with the same settings.

\subsection{Compute Resources}
Evaluating PEARL without training on PSVU-Bench takes approximately 20 minutes on 8 NVIDIA H200 GPUs. Constructing PSVU-10K runs Qwen3.8-Flash-Next with vLLM on up to six 8-GPU nodes. Training with LoRA on 8 NVIDIA H200 GPUs takes about 11 minutes for the retriever and 4.9 hours for the VLM, and the trained PEARL runs the same pipeline.

\section{Annotation and Quality Control}
\label{app:annotation-quality}
\begin{figure}[H]
\centering
\includegraphics[width=\linewidth]{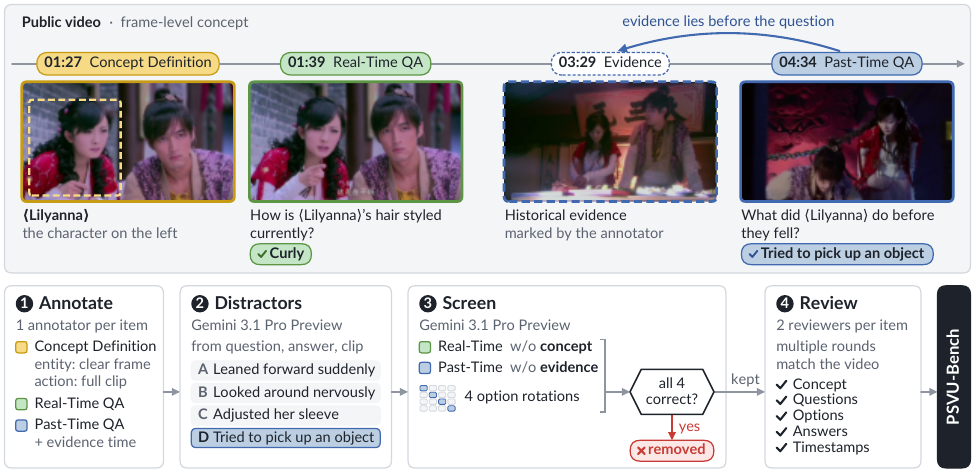}
\caption{\textbf{PSVU-Bench annotation and quality control.} A released video example illustrates concept and QA annotation, distractor generation, ablation-based filtering, and human review.}
\label{fig:annotation-pipeline}
\end{figure}

Figure~\ref{fig:annotation-pipeline} illustrates the annotation and quality-control workflow with an example from the released data.

\noindent\textbf{Annotation.} Six annotators define concepts when the target appears and write Real-Time and Past-Time questions at later timestamps. For entity concepts, they select frames where the target is clearly visible. For action concepts, they manually annotate clip boundaries to include the complete action. For Past-Time QA, they also identify the historical evidence timestamp. Gemini 3.1 Pro Preview generates multiple-choice distractors from the annotators' questions, correct answers, and corresponding video clips. Question templates are provided in Sec.~\ref{app:templates}.

\noindent\textbf{Automated filtering.} We use Gemini 3.1 Pro Preview with the original evaluation prompt, which presents the question and options and requests the answer choice in \texttt{<ans>} tags. Real-Time QA is evaluated with and without the defined concepts, and Past-Time QA with and without historical evidence clips. Each ablated question is evaluated under four option rotations (Sec.~\ref{app:option-rotation}). A question is filtered if the model answers all four correctly without the relevant information, regardless of its answer with full information.

\noindent\textbf{Human verification.} Each QA item is annotated by one of the six annotators and reviewed by two of the four reviewers, who conduct multiple rounds of inspection to check that QA items and timestamps align with the video. Human evaluation scores are reported in Table~\ref{tab:main-table}.

\section{PSVU-10K Construction}
\label{app:psvu10k}
\begin{figure}[H]
\centering
\includegraphics[width=\linewidth]{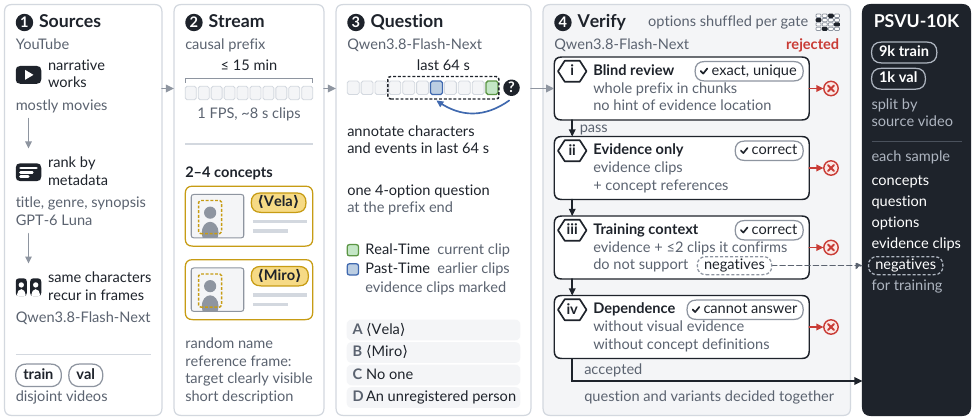}
\caption{\textbf{PSVU-10K construction.} Qwen3.8-Flash-Next writes one Real-Time or Past-Time question at the end of each causal stream prefix, and a question is kept only if it passes four verification gates run with shuffled options. Gate (iii) also provides the verified negative clips used for training.}
\label{fig:psvu10k-pipeline}
\end{figure}

Figure~\ref{fig:psvu10k-pipeline} summarizes how PSVU-10K is constructed.

\noindent\textbf{Sources.} PSVU-10K is built from YouTube videos of 155 narrative works, mostly movies. We rank candidate works by their metadata with GPT-6 Luna and keep those in which the same characters recur, which the open-source Qwen3.8-Flash-Next confirms by inspecting frames. Training and validation questions come from disjoint videos. Each video's metadata also names the work it comes from, and we drop every candidate work that appears in PSVU-Bench, together with its other uploads, seasons and episodes. PSVU-10K therefore shares no work with the benchmark, so training never sees the characters on which PEARL is evaluated.

\noindent\textbf{Streams and concepts.} Each sample is a causal prefix of a video with a median length of 7.5 minutes (from 1 to 15 minutes), sampled at 1 FPS and split into clips of about 8 seconds. It registers two to four concepts (median four), each with a randomly sampled name, a reference frame in which the target is clearly visible, and a short description.

\noindent\textbf{Question generation.} Qwen3.8-Flash-Next annotates the characters and events in the last 64 seconds of each prefix and writes one four-option question at its end. Real-Time questions ask about the current clip, and Past-Time questions ask about earlier evidence, which lies a median of 32 seconds and up to 6.5 minutes before the question. The questions cover actions, relations between characters, the recipients of actions, the order of events, locations and objects. Options include the other registered names and distractors such as ``No one'' or ``An unregistered person''. Action concepts are not included.

\noindent\textbf{Verification.} A question is accepted only if it passes all of the following checks, each run with shuffled options: (i) after reviewing the whole prefix in chunks, without any hint of the evidence location, Qwen3.8-Flash-Next selects exactly the annotated answer; (ii) it answers correctly from the evidence clips and the concept references alone; (iii) it answers correctly from the training context, namely the evidence clips plus at most two clips that it has confirmed not to support the answer; (iv) it cannot answer once the visual evidence is withheld, or once the concept definitions are removed.

\noindent\textbf{Statistics.} PSVU-10K contains 10k questions: 9k for training (3,616 Real-Time and 5,384 Past-Time) and 1k for validation (359 Real-Time and 641 Past-Time).

\section{Data Release and License}
\label{app:data-release}
We publicly release PSVU-Bench, PSVU-10K, the evaluation and training code, the data construction scripts and the trained LoRA weights for research use. The code and scripts are released under the MIT License, while the annotations, metadata, and rendered synthetic videos created by us are released under CC BY-NC 4.0. The trained LoRA weights are released under the Apache 2.0 License of their base models. For YouTube videos in PSVU-Bench and PSVU-10K, we will not redistribute raw video files, and will instead release video IDs and timestamps so that users can obtain the videos from YouTube under its terms of service. The character models and motions of the synthetic Action split are generated with GPT-6 Astra, then arranged and annotated by hand. We also provide a dataset card covering sources, intended use, license notes, privacy considerations, and a removal request procedure.

\section{Additional Experimental Results}

\subsection{Training Other Base Models}
\label{app:train-other-vlms}
To check that the benefit of PSVU-10K is not specific to Qwen3-VL-8B, we apply the same training to LLaVA-OV-7B+PEARL and Qwen2-VL-7B+PEARL. Both reuse the trained retriever, which does not depend on the answering VLM, and only their answering VLMs are fine-tuned with LoRA, using the same hyperparameters and number of steps as Qwen3-VL-8B (Sec.~\ref{app:training-settings}).

\begin{table}[htbp]
\centering
\caption{\textbf{Training across base models} on PSVU-Bench. $^\dagger$ denotes training on PSVU-10K, with gains over the same model without training marked by arrows. LLaVA-OV-7B and Qwen2-VL-7B reuse the trained retriever and train only their answering VLMs.}
\label{tab:train-other-vlms}
\setlength{\tabcolsep}{4pt}
\resizebox{\textwidth}{!}{%
\begin{tabular}{@{} l|cccccc c@{\hspace{2.4em}}|cc c@{\hspace{2.4em}}|c@{\hspace{2.4em}} @{}}
\toprule
\multirow{2}{*}{Method} & \multicolumn{7}{c|}{Real-Time QA} & \multicolumn{3}{c|}{Past-Time QA} & \multirow{2}{*}{Overall Avg} \\ \cmidrule(lr){2-8} \cmidrule(lr){9-11}
& Presence & Behavior & Appearance & Location & Relation & Action & Avg & Event & Time & Avg & \\ \midrule
LLaVA-OV-7B+PEARL & 35.05 & 32.61 & 34.24 & 31.34 & 34.86 & 20.06 & 31.36 & 45.08 & 40.30 & 42.69 & 37.03 \\
LLaVA-OV-7B+PEARL$^\dagger$ & 40.84 & 43.48 & 40.22 & 42.54 & 43.12 & 22.56 & 38.79\rlap{\,\textcolor{teal}{\tiny$\uparrow$7.43}} & 54.92 & 47.76 & 51.34\rlap{\,\textcolor{teal}{\tiny$\uparrow$8.65}} & 45.07\rlap{\,\textcolor{teal}{\tiny$\uparrow$8.04}} \\
\midrule
Qwen2-VL-7B+PEARL & 35.37 & 32.07 & 33.70 & 31.34 & 33.94 & 24.23 & 31.78 & 47.15 & 41.79 & 44.47 & 38.13 \\
Qwen2-VL-7B+PEARL$^\dagger$ & 42.12 & 41.30 & 44.02 & 41.04 & 44.95 & 27.58 & 40.17\rlap{\,\textcolor{teal}{\tiny$\uparrow$8.39}} & 55.44 & 52.24 & 53.84\rlap{\,\textcolor{teal}{\tiny$\uparrow$9.37}} & 47.01\rlap{\,\textcolor{teal}{\tiny$\uparrow$8.88}} \\
\midrule
Qwen3-VL-8B+PEARL & 63.67 & 50.00 & 55.43 & 52.99 & 53.21 & 48.47 & 53.96 & 52.85 & 48.76 & 50.80 & 52.38 \\
Qwen3-VL-8B+PEARL$^\dagger$ & 68.49 & 62.50 & 62.50 & 65.67 & 64.22 & 52.37 & 62.62\rlap{\,\textcolor{teal}{\tiny$\uparrow$8.66}} & 62.18 & 58.21 & 60.19\rlap{\,\textcolor{teal}{\tiny$\uparrow$9.39}} & 61.41\rlap{\,\textcolor{teal}{\tiny$\uparrow$9.03}} \\ \bottomrule
\end{tabular}%
}
\end{table}

As shown in Table~\ref{tab:train-other-vlms}, training improves every category for all three base models. Real-Time accuracy rises by 7.43\% to 8.66\% and Past-Time accuracy by 8.65\% to 9.39\%, and Action accuracy rises by 2.50\% to 3.90\% although PSVU-10K contains no action concepts. These similar gains across three architectures show that the benefit of PSVU-10K is not tied to one VLM, and a single trained retriever serves all of them.

\subsection{Historical Evidence Retrieval}
\label{app:evidence-retrieval}
We evaluate retrieval on frame-level Past-Time questions whose annotated evidence timestamp lies in an eligible historical clip. A hit requires a returned clip to contain that timestamp, using a start-inclusive, end-exclusive interval. The metric is conservative, because clips that support the answer at other moments are not counted.

On average, a complete video contains 456.6 clips, including 241.4 eligible historical candidates at the question time. We compare random selection and PEARL's ranking at several selection budgets. For the default setting, both select up to four candidates and add one adjacent clip on either side, followed by deduplication and temporal filtering. The random baseline selects candidates uniformly without replacement, and its results are analytical expectations.

\begin{table}[htbp]
\centering
\caption{Timestamp-based historical evidence retrieval at different selection budgets. Avg. clips counts selected historical clips, random results are analytical expectations, and Trained denotes PEARL with the retriever trained on PSVU-10K.}
\label{tab:evidence-retrieval}
\begin{tabular}{lrrrrrr}
\toprule
& \multicolumn{3}{c}{Avg. clips} & \multicolumn{3}{c}{Evidence hit rate (\%)} \\
\cmidrule(lr){2-4} \cmidrule(lr){5-7}
Selection & Random & PEARL & Trained & Random & PEARL & Trained \\ \midrule
Top-1 & 1.00 & 1.00 & 1.00 & 0.80 & 7.85 & \textbf{14.66} \\
Top-2 & 2.00 & 2.00 & 2.00 & 1.60 & 13.87 & \textbf{23.82} \\
Top-4 & 3.99 & 3.99 & 3.99 & 2.94 & 22.25 & \textbf{32.46} \\
Top-4 + neighbors & 11.68 & 10.76 & 10.56 & 7.90 & 34.82 & \textbf{44.50} \\ \bottomrule
\end{tabular}
\end{table}

Table~\ref{tab:evidence-retrieval} shows that PEARL outperforms random selection at every budget, and adjacent-clip expansion adds 12.57\% to its evidence hit rate. Training the retriever on PSVU-10K raises the hit rate at every budget, by 6.81\% at Top-1 and 9.68\% with expansion, while selecting slightly fewer clips after expansion.

\subsection{Real-World Interaction: Delayed and Ambiguous Instructions}
\label{app:interaction-robustness}
PSVU-Bench registers each concept at a moment when its target is clearly visible. In live use, however, a user may give the instruction a few seconds after the target has left the frame, or refer to the target ambiguously while several candidates are visible. In both cases the current frame is an unreliable reference, and a wrong reference misleads every later question about the concept. We therefore test two training-free extensions of PEARL: a short causal buffer for delayed instructions and selective clarification for ambiguous ones.

We evaluate Qwen3-VL-8B+PEARL on 50 entity registrations and 241 associated registration--question pairs. For instructions delayed by 2, 4, or 8 seconds, Qwen3-VL-8B selects a reference frame from a causal 8-second buffer sampled at 1 FPS without access to the original registration time. For ambiguous instructions such as \textit{``this character''}, the model may request one clarification, answered with a scripted description of the intended entity. A binding is correct if the reference image and description stored in Concept Memory match the intended entity. The clarification prompt was developed on these cases.

\begin{table}[htbp]
\centering
\caption{Mean QA accuracy (\%) on the controlled interaction cases, averaging Real-Time and Past-Time question accuracies equally. For delayed instructions, the baseline uses the arrival frame and the extension uses the buffer. For ambiguous instructions, they use direct registration and selective clarification, respectively.}
\label{tab:interaction-robustness}
\begin{tabular}{lcc}
\toprule
Interaction & Baseline & Extension \\ \midrule
Aligned, clear & \multicolumn{2}{c}{45.39} \\
2-second delay & 42.41 & \textbf{44.07} \\
4-second delay & 40.85 & \textbf{44.18} \\
8-second delay & 42.16 & \textbf{44.73} \\
Ambiguous instruction & 37.05 & \textbf{46.15} \\ \bottomrule
\end{tabular}
\end{table}

As shown in Table~\ref{tab:interaction-robustness}, buffering raises QA accuracy by 1.66\%, 3.33\% and 2.57\% at delays of 2, 4 and 8 seconds, and binds all 50 targets correctly at every delay. Selective clarification raises accuracy on ambiguous instructions by 9.10\%, on par with aligned, clear instructions. It requests clarification on 30 of the 50 ambiguous inputs and on none of the 50 clear controls, and binds all 30 clarified targets and 44 of the 50 targets overall correctly. Both extensions change only concept registration and leave the rest of PEARL unchanged, so they can be added whenever delayed or ambiguous definitions are expected.

\subsection{Hyperparameter Analysis for Past-Time QA}
\begin{figure}[ht]
  \centering
  \includegraphics[width=0.6\linewidth]{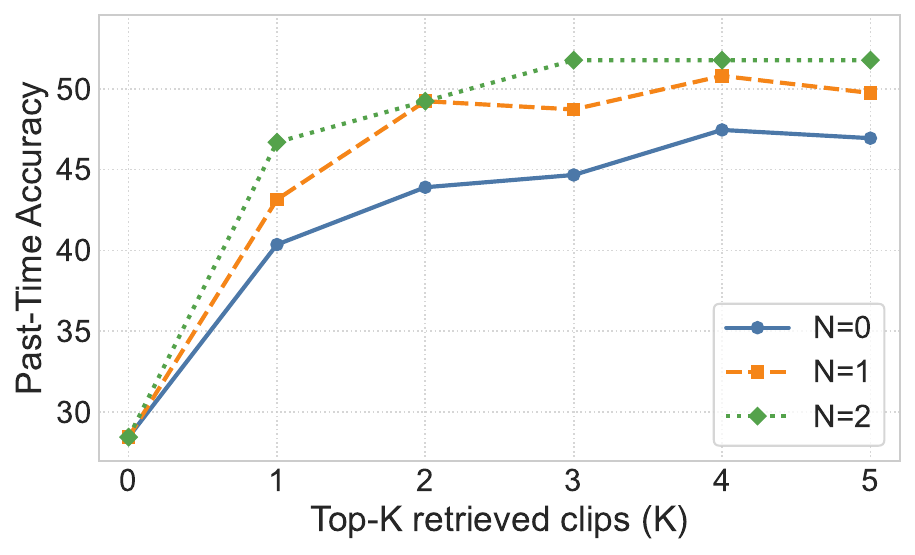}
  \caption{Past-Time accuracy under different top-$K$ ($K$) and expansion sizes ($N$).}
  \label{fig:Hyperparameter1}
\end{figure}

Past-Time QA depends on how much history PEARL passes to the VLM. This amount is set by two hyperparameters, the number of retrieved clips $K$ and the number of adjacent clips $N$ added on each side of a retrieved clip, and larger values cover more evidence at the cost of a longer input. As shown in Fig.~\ref{fig:Hyperparameter1}, a single retrieved clip raises accuracy by 12\% to 18\% over no retrieval ($K=0$), and the gains slow down beyond $K=2$. Adjacent clips improve accuracy at every $K>0$. At $K=4$, $N=1$ adds about 3\% over $N=0$, while $N=2$ adds only about 1\% more and increases the input from at most 12 to at most 20 clips. We therefore adopt $K=4$ and $N=1$ as the default configuration.

\subsection{Hyperparameter Analysis for Real-Time QA}
\begin{figure}[ht]
  \centering
  \includegraphics[width=0.6\linewidth]{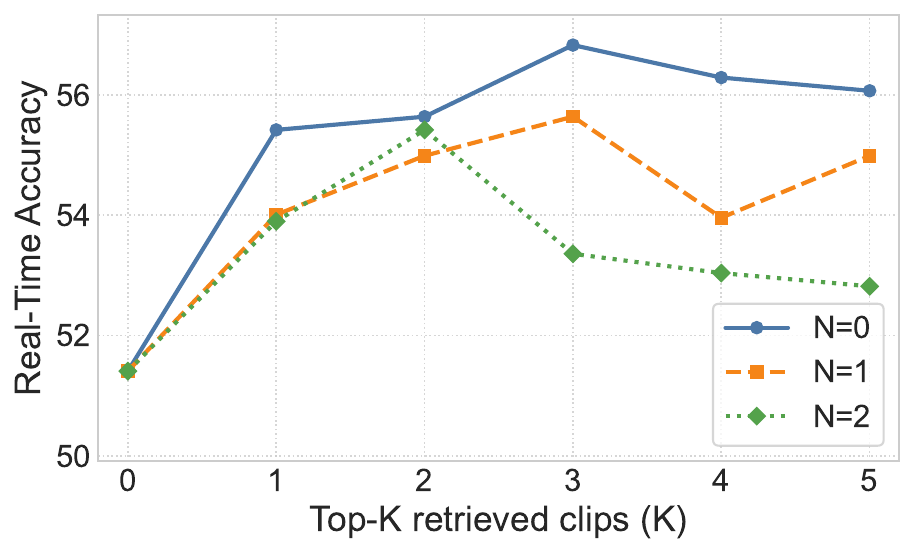} 
  \caption{Real-Time accuracy under different top-$K$ ($K$) and expansion sizes ($N$).}
  \label{fig:Hyperparameter_realtime}
\end{figure}

PEARL does not know in advance whether a question concerns the present or the past, so it also retrieves historical clips for Real-Time questions. We therefore examine whether this history helps or distracts Real-Time QA. As shown in Fig.~\ref{fig:Hyperparameter_realtime}, every setting with retrieval outperforms no retrieval ($K=0$), and a single retrieved clip already adds 2.49\% to 4.01\%. We attribute this gain to concept-aware retrieval. The rewritten query describes the queried concept, so the retrieved clips show the concept at other moments and add views of its appearance beyond the single reference frame. Without query rewriting, the same retrieval returns clips unrelated to the concept and lowers Real-Time accuracy by 1.19\% (Table~\ref{tab:my-table}). For $K\geq3$, accuracy decreases as $N$ grows, because adjacent clips add history that is unrelated to the queried concept. Across all settings with retrieval, accuracy stays between 52.82\% and 56.83\%, so Real-Time QA is robust to $K$ and $N$. The default $K=4$ and $N=1$, chosen for Past-Time QA, remains within 3\% of the best Real-Time accuracy.

\subsection{Effect of Model Scales}
PEARL is designed to work with off-the-shelf VLMs, so we test whether its gains hold across model scales. We evaluate the Qwen2-VL series~\citep[2B and 7B][]{wang2024qwen2} and the Qwen3-VL series~\citep[4B and 8B][]{bai2025qwen3} with and without PEARL on PSVU-Bench (Table~\ref{tab:model-scales}), and focus on models up to 8B because PSVU requires real-time responses.

\begin{table}[ht]
\centering
\caption{Performance comparison across different model scales on PSVU-Bench. \textbf{Bold} and \underline{underline} denote the best and second-best results within each model family (Qwen2-VL and Qwen3-VL), respectively.}
\label{tab:model-scales}
\resizebox{0.75\columnwidth}{!}{%
\begin{tabular}{@{} l | wc{2.2cm} wc{2.3cm} wc{2.3cm} @{}}
\toprule
Model & Real-Time & Past-Time & Avg \\ \midrule
Qwen2-VL-2B~\citep{wang2024qwen2} & \underline{31.24} & 22.84 & 27.04 \\
Qwen2-VL-7B~\citep{wang2024qwen2} & 22.29 & \underline{35.84} & 29.07 \\
\textbf{Qwen2-VL-2B+PEARL} & 29.93\rlap{\,\textcolor{red}{\tiny$\downarrow$1.31}} & 32.49\rlap{\,\textcolor{teal}{\tiny$\uparrow$9.65}} & \underline{31.21}\rlap{\,\textcolor{teal}{\tiny$\uparrow$4.17}} \\
\textbf{Qwen2-VL-7B+PEARL} & \textbf{31.78}\rlap{\,\textcolor{teal}{\tiny$\uparrow$9.49}} & \textbf{44.47}\rlap{\,\textcolor{teal}{\tiny$\uparrow$8.63}} & \textbf{38.13}\rlap{\,\textcolor{teal}{\tiny$\uparrow$9.06}} \\ \midrule
Qwen3-VL-4B~\citep{bai2025qwen3} & 24.08 & 30.96 & 27.52 \\
Qwen3-VL-8B~\citep{bai2025qwen3} & 27.16 & 30.25 & 28.71 \\
\textbf{Qwen3-VL-4B+PEARL} & \underline{40.78}\rlap{\,\textcolor{teal}{\tiny$\uparrow$16.70}} & \underline{50.25}\rlap{\,\textcolor{teal}{\tiny$\uparrow$19.29}} & \underline{45.52}\rlap{\,\textcolor{teal}{\tiny$\uparrow$18.00}} \\
\textbf{Qwen3-VL-8B+PEARL} & \textbf{53.96}\rlap{\,\textcolor{teal}{\tiny$\uparrow$26.80}} & \textbf{50.80}\rlap{\,\textcolor{teal}{\tiny$\uparrow$20.55}} & \textbf{52.38}\rlap{\,\textcolor{teal}{\tiny$\uparrow$23.67}} \\ \bottomrule
\end{tabular}%
}
\end{table}

\noindent\textbf{Gains across model scales.} PEARL raises the average accuracy of all four models, and its gain grows with model size, from 4.17\% to 9.06\% within Qwen2-VL and from 18.00\% to 23.67\% within Qwen3-VL. Past-Time accuracy rises for every model, by 8.63\% to 20.55\%.

\noindent\textbf{Scaling with PEARL.} Without PEARL, a larger model gains little: the average rises by 2.03\% from Qwen2-VL-2B to 7B and by 1.19\% from Qwen3-VL-4B to 8B, since a larger model cannot make up for the concept references and historical evidence that PEARL supplies. With PEARL, the same scaling adds 6.92\% and 6.86\%, so PEARL lets a larger model turn its capacity into accuracy on PSVU.

\subsection{Generalization on Non-personalized Tasks}
\label{app:general-streaming}
PEARL is designed for user-defined concepts, so we check that it does not harm general streaming video understanding. We evaluate the base models with and without PEARL on the non-personalized OVO-Bench~\citep{niu2025ovo}. As shown in Table~\ref{tab:general-streaming}, PEARL raises the average accuracy by 2.60\% for LLaVA-OV-7B and 2.55\% for Qwen2-VL-7B, with gains on both Real-Time and Backward questions. PEARL therefore adds personalization without compromising general streaming ability.

\begin{table}[t]
\centering
\caption{Comparison of base models with and without PEARL on OVO-Bench.}
\label{tab:general-streaming}
\resizebox{0.5\linewidth}{!}{%
\begin{tabular}{@{}cccc@{}}
\toprule
Model                      & Real-Time & Backward & Avg   \\ \midrule
LLaVA-OV-7B                & 64.02     & 43.71    & 53.87 \\
\textbf{LLaVA-OV-7B+PEARL} & 65.34     & 47.60    & 56.47 \\ \midrule
Qwen2-VL-7B                & 55.98     & 46.46    & 51.22 \\
\textbf{Qwen2-VL-7B+PEARL} & 58.02     & 49.51    & 53.77 \\ \bottomrule
\end{tabular}%
}
\end{table}

\end{document}